\documentclass{article} 
\usepackage{iclr2025_conference,times}

\usepackage{hyperref}[citecolor=cyan]
\hypersetup{
    colorlinks = true,
    citecolor = {cyan},
}

\usepackage{amsfonts,bm}
\usepackage[most,skins,theorems]{tcolorbox}

\tcbset{
  aibox/.style={
    width=\linewidth,
    top=8pt,
    bottom=4pt,
    colback=blue!6!white,
    colframe=black,
    colbacktitle=black,
    enhanced,
    center,
    attach boxed title to top left={yshift=-0.1in,xshift=0.15in},
    boxed title style={boxrule=0pt,colframe=white,},
  }
}
\usepackage{hyperref}[citecolor=magenta]

\newcommand{\forceindent}{\leavevmode{\parindent=1em\indent}}

\newcommand{\model}{PTR}
\newcommand{\PR}{progressive refinement}

\usepackage{arydshln}
\usepackage{array}
\usepackage{amsmath}
\usepackage{amssymb}
\usepackage{booktabs}
\usepackage{colortbl}
\usepackage{etoolbox}
\usepackage{enumitem}
\usepackage{makecell}
\usepackage{multirow}
\usepackage{xspace}
\usepackage{graphicx}
\usepackage{spverbatim}
\AtBeginEnvironment{spverbatim}{\small}

\definecolor{Ground}{RGB}{255,184,55}
\definecolor{Rice}{RGB}{251,248,238}
\definecolor{Dirt}{RGB}{191,169,115}
\definecolor{Pink}{RGB}{226,184,176}
\definecolor{Violet}{RGB}{163,148,170}
\definecolor{mygray}{RGB}{226, 226, 226}
\definecolor{Green}{RGB}{144, 238, 144}
\definecolor{Gray}{RGB}{211, 211, 211}
\definecolor{Blue}{RGB}{173, 216, 230}

\newcolumntype{a}{>{\columncolor{Green!10}}c}
\newcolumntype{e}{>{\columncolor{Blue!10}}c}
\newcolumntype{g}{>{\columncolor{Ground!9}}c}
\newcolumntype{d}{>{\columncolor{Dirt!10}}c}
\newcolumntype{f}{>{\columncolor{Pink!10}}c}
\newcolumntype{v}{>{\columncolor{Violet!9}}c}
\newcolumntype{P}[1]{>{\centering\arraybackslash}p{#1}}

\def\eqref#1{equation~\ref{#1}}

\def\1{\bm{1}}

\DeclareMathAlphabet{\mathsfit}{\encodingdefault}{\sfdefault}{m}{sl}
\SetMathAlphabet{\mathsfit}{bold}{\encodingdefault}{\sfdefault}{bx}{n}

\definecolor{blanchedalmond}{rgb}{1.0, 0.92, 0.8}
\definecolor{carmine}{rgb}{0.59, 0.0, 0.09}
\definecolor{lightblue}{rgb}{0.22,0.45,0.70}

\renewcommand{\mathbf}{\boldsymbol}

\makeatletter
\def\Ddots{\mathinner{\mkern1mu\raise\p@
\vbox{\kern7\p@\hbox{.}}\mkern2mu
\raise4\p@\hbox{.}\mkern2mu\raise7\p@\hbox{.}\mkern1mu}}
\makeatother

\numberwithin{equation}{section}

\definecolor{amaranth}{rgb}{0.9, 0.17, 0.31}
\definecolor{antiquebrass}{rgb}{0.8, 0.58, 0.46}
\definecolor{antiquefuchsia}{rgb}{0.57, 0.36, 0.51}
\definecolor{chromeyellow}{rgb}{0.31, 0.47, 0.26}

\definecolor{Bluelite}{HTML}{2b8cbe}
\definecolor{Bluedeep}{HTML}{084081}

\definecolor{Orglite}{HTML}{d95f0e}
\definecolor{Orgdeep}{HTML}{662506}

\definecolor{PeriIte}{HTML}{6a51a3}
\definecolor{Perdeep}{HTML}{3f007d}

\definecolor{Brwonlite}{HTML}{d95f0e}
\definecolor{Browndeep}{HTML}{662506}

\definecolor{DeepBlue}{HTML}{6a51a3}
\definecolor{DeepBlueDeep}{HTML}{3f007d}

\usepackage{tikz}  

\usepackage{algorithm}
\usepackage[noend]{algpseudocode}
\usepackage[utf8]{inputenc} %
\usepackage[T1]{fontenc}    %
\usepackage{url}            %
\usepackage{booktabs}       %
\usepackage{amsfonts}       %
\usepackage{nicefrac}       %
\usepackage{microtype}      %
\usepackage{xcolor}         %
\usepackage{subcaption}
\usepackage{array}
\usepackage{arydshln}
\usepackage{wrapfig}
\usepackage{fdsymbol}
\usepackage{listings}
\usepackage{caption}

\usepackage{amsmath}
\usepackage{graphicx}
\usepackage{amsmath}
\usepackage{amssymb}
\usepackage{booktabs}
\usepackage{multirow}
\usepackage{diagbox}
\usepackage{bm}
\usepackage{fleqn, tabularx}
\usepackage{bbm}
\usepackage{mathrsfs}
\usepackage{enumitem}
\usepackage{tablefootnote}
\usepackage{listings}
\usepackage{array}
\usepackage{wrapfig}
\definecolor{gred}{RGB}{219,68,55}
\definecolor{ggreen}{RGB}{15,157,88}
\usepackage{epigraph}
\newtcolorbox{AIbox}[2][]{aibox,title=#2,#1}
\definecolor{lightblue}{rgb}{0.22,0.45,0.70}%
\DeclareCaptionFont{white}{\color{white}}
\DeclareCaptionFormat{listing}{%
  \parbox{\textwidth}{\colorbox{gray}{\parbox{\textwidth}{#1#2#3}}\vskip-4pt}}
\usepackage{url}

\title{Think Thrice Before You Act: Progressive Thought Refinement in Large Language Models}

\author{
    \small Chengyu Du$^{14}$, Jinyi Han$^{2}$, Yizhou Ying$^{1}$, Aili Chen$^{14}$, Qianyu He$^{1}$, Haokun Zhao$^{1}$, \\
    \small \textbf{Sirui Xia$^{14}$, Haoran Guo$^{5}$, Jiaqing Liang$^{3}$, Zulong Chen$^{4}$, Liangyue Li$^{4}$, Yanghua Xiao$^{*1}$} \\ 
    \small $^1$Shanghai Key Laboratory of Data Science, School of Computer Science, Fudan University \\
    \small $^2$Shanghai Institute of Artificial Intelligence for Education, East China Normal University \\
    \small $^3$School of Data Science, Fudan University  \quad
    \small $^4$ Alibaba Group \quad
    \small $^5$RhineAI \\
    \small \texttt{\{cydu24, yzying24, alchen24, qyhe21, hkzhao23, srxia24\}@m.fudan.edu.cn}, \\
    \small \texttt{\{liangjiaqin, shawyh\}@fudan.edu.cn}, 
    \small \texttt{\{jinyihan099, rhineailab\}@gmail.com}, \\
    \small \texttt{\{zulong.czl, liangyue.lly\}@alibaba-inc.com}
}

\iclrfinalcopy
\begin{document}

\maketitle
\begin{abstract}
  Recent advancements in large language models (LLMs) have demonstrated that progressive refinement, 
  rather than providing a single answer, results in more accurate and thoughtful outputs. 
  However, existing methods often rely heavily on supervision signals to evaluate previous responses, making it difficult to effectively assess output quality in more open-ended scenarios. Additionally, these methods are typically designed for specific tasks, which limits their generalization to new domains.
  To address these limitations, we propose Progressive Thought Refinement (PTR), a framework that enables LLMs to progressively refine their responses. PTR operates in two phases: 
  (1) Thought data construction stage: We propose a \textit{weak and strong model collaborative selection} strategy to build a high-quality progressive refinement dataset to ensure logical consistency from thought to answers, and the answers are gradually refined in each round.
  (2) Thought-Mask Fine-Tuning Phase: 
        We design a training structure to mask the "thought" and adjust loss weights to encourage LLMs to refine prior thought, teaching them to implicitly understand "how to improve" rather than "what is correct."
        Experimental results show that PTR significantly enhances LLM performance across ten diverse tasks (avg. from 49.6\% to 53.5\%) without task-specific fine-tuning. 
        Notably, in more open-ended tasks, LLMs also demonstrate substantial improvements in the quality of responses beyond mere accuracy, suggesting that PTR truly teaches LLMs to self-improve over time.
\end{abstract}

\section{Introduction} 

\begin{flushright}
  \begin{minipage}{1.2\textwidth}  
    \begin{quote}
      ``\textit{Think thrice before you act.}'' 
    \end{quote}
  \end{minipage}
\end{flushright}
\begin{flushright}
  \begin{minipage}{0.6\textwidth} 
    \begin{quote}
        \hfill — \underline{Confucius}  
    \end{quote}
  \end{minipage}
\end{flushright}

Recent advancements in large language models (LLMs) have highlighted that
\PR{} is more important than simply providing a single answer~\citep{yang2023leandojo,madaan2023self}.
Humans often rely on a combination of two thinking systems to solve problems, 
known as \textit{System 1} and \textit{System 2}~\citep{kahneman2011thinking}.
\textit{System 1} facilitates quick, intuitive responses 
but often lacks the depth required to handle complex reasoning tasks.
In contrast, \textit{System 2} engages in progressive refinement, 
gradually improving a solution by starting with a rough approximate thought
and iteratively adding detail and accuracy. 
Recent work, such as GPT-o1~\citep{openai2024gpto1}, 
demonstrates that LLMs perform better by adopting progressive thought refinement.
This approach leads to more accurate and thoughtfully considered outcomes, 
similar to how the human brain processes complex tasks.

Progressive refinement ability is imperative for LLMs because it significantly enhances the quality 
of responses by gradually improving accuracy and depth. 
Previous methods heavily rely on supervision signals, such as correctness assessments, to assess response quality.
For example, labeled datasets with feedback are used to fine-tune models 
as verifiers~\citep{han2024smalllanguagemodelselfcorrect,havrilla2024glore,welleck2023generating}, 
facilitating self-assessment and iterative improvement. 
Additionally, Reinforcement Learning (RL) reward functions are also 
employed to guide models toward generating better answers~\citep{chen2024self,yuan2024self,rosset2024direct,akyurek-etal-2023-rl4f}.
However, evaluating answers based on supervision signals has limitations, 
as annotators often \textbf{struggle to provide accurate labels} without clear, comprehensive criteria. 
This is particularly challenging in open-ended tasks, such as text generation and summarizing, 
where the distinction between ``correct'' and ``incorrect'' is blurred, 
making it difficult to define and evaluate response quality.

Due to significant variations in supervision signals and evaluation criteria across tasks,
previous self-improvement approaches have primarily aimed to enhance accuracy within specific domains.
Examples include enabling LLMs to self-debug for improved code generation~\citep{chen2023teaching, tony2024promptingtechniquessecurecode, liang2023code} 
and solving math problems through progressive step validation~\citep{wang2023math, lightman2023let, uesato2022solving}.
These methods often rely on task-specific pipelines or reward models, making generalization difficult.
The key limitation is that errors addressed in one domain may not apply to other tasks, since different tasks exhibit varying error types.
Consequently, transferring these approaches to new tasks often fails~\citep{tian2024selfimprovementllmsimaginationsearching}
, and models trained with these methods have \textbf{limited generalization capabilities}, struggling to improve performance beyond their training domains.

To address these challenges, we introduce \model{} (\textbf{P}rogressive \textbf{T}hought \textbf{R}efinement), 
a framework specifically designed to stimulate the model's intrinsic refinement ability. 
Our \model{} method comprises a \PR{} dataset construction phase and a weighted thought-mask fine-tuning phase. 
During the \PR{} dataset construction phase, we obtain queries from open-domain datasets and employ 
a \textit{weak-strong model collaborative selection} strategy to construct 
high-quality \textit{thoughts} and \textit{refined answers} dataset. 
This strategy not only ensures improvement from thoughts to answers but also \textbf{eliminates the need for accurate labels}. 
In the fine-tuning phase, we employ \textit{weighted thought-mask fine-tuning} to teach LLMs 
to implicitly understand ``how to improve'' rather than supervising them with ``what is correct''. 
Specifically, we reformulate the masked data structure and redesign the 
loss of weighting to encourage LLMs to improve responses based on previous thoughts
and ensuring logical consistency between the thought process and the final answer.

Our experimental results show that LLMs trained with \model{} can improve the quality of their previous answers 
across ten tasks, including knowledge reasoning, code generation, mathematical reasoning, comprehension, 
summarizing, and text generation. The average performance across these tasks improved from 49.6\% to 53.5\%, 
with a significant improvement on the MMLU task, where accuracy increased from 57.1\% to 64.1\% for Qwen2-8B. 
Notably, these improvements occur \textbf{without task-specific fine-tuning}, demonstrating that our method activates 
the model to learn \PR{} from the PTR dataset. Moreover, in more open-ended tasks, 
LLMs have also demonstrated further improvements in answer quality and formatting beyond correctness.

Our contributions are threefold: 
\begin{itemize} 
  \item We propose the \model{} method to stimulate models' \PR{} abilities and enhance generalization across various tasks without additional task-specific fine-tuning. 
  \item We design an efficient \textit{weak-strong model collaborative selection} strategy to construct high-quality PTR datasets without extra feedback. 
  \item We introduce a novel \textit{weighted thought-mask fine-tuning} method to instill general \PR{} capabilities in LLMs. 
\end{itemize}

\section{related work}

\begin{figure}[t]
  \centering
  \includegraphics[width=0.98\linewidth]{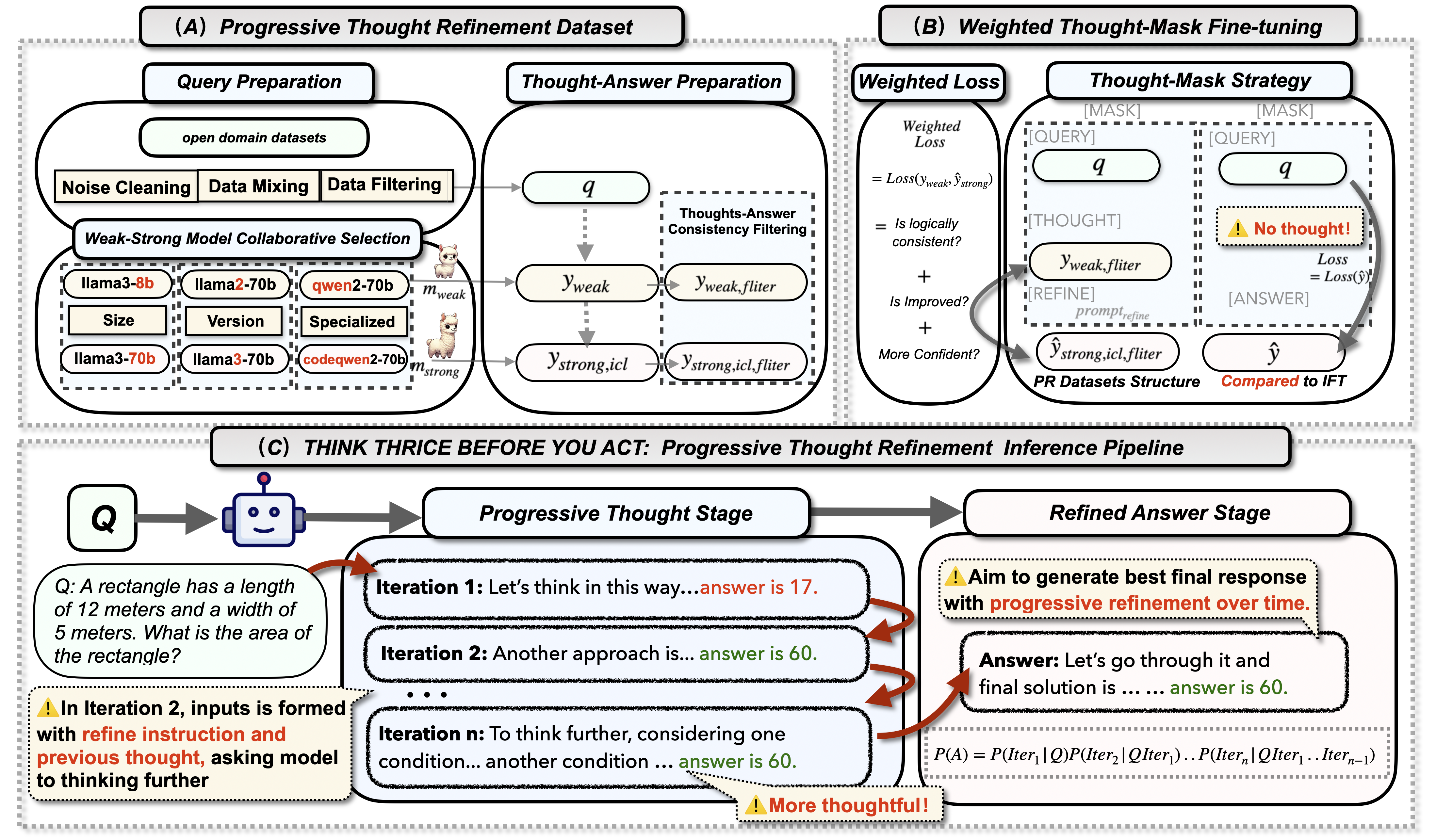}
 
  \caption{Illustration Our approaches. (A) Pipeline of our \PR{} Dataset construction. We first prepare queries from the general open domain datasets, and pre-processing queries in three steps.
  Then we use a strong weak model collaborative selection strategy to generate thoughts and answers for each query.
  We also implement In-context Learning (ICL) and Consistency Filtering to ensure the quality of the thought process.
  (B) The illustration of Weighted Thought Masking Fine-tuning. Aiming at training the model to produce a better response 
  in the next attempt and ensure logical consistency during the thought process.  The difference between our method and IFT is that we use thought-mask techniques to ask model to generate better responses.
  (C) Pipeline of our \model{}. Given a query $Q$, 
  LLMs think progressively and refine their responses based on their own previous thought and refinement instruction.
  LLMs refined its mistakes on the second attempt, as well as gave a more thoughtful answer at a later iteration. 
}
 
  \label{fig:main}
\end{figure}

\textbf{Progressive Refinement with External Feedback}
Existing work often relies on external tools or stronger LLMs to provide feedback for refinement. 
For example, external tools are used to critique and provide feedback on the primary model's 
responses~\citep{yang2023auto, chen2023teaching, charalambous2023new, nijkamp2022codegen, yao2022react, gou2023critic}. 
Models have improved their code generation capabilities by leveraging error messages from the Python interpreter~\citep{wang2023leti} and by  teaching large language models to debug and explain their own code, allowing them to identify and fix errors without human feedback~\citep{chen2023teaching}. Similarly, compiler feedback has been utilized in code generation~\citep{chen2024self, olausson2023self}. Additionally, some approaches utilize criticisms or constraints generated by stronger models~\citep{pan2023automatically, du2023improving, bai2022constitutional, huang2023agentcoder}, such as using a strong model to verify the correctness of another model's math solutions~\citep{wang2023math}, thereby relying on external information sources to guide improvements.
Although models can self-correct through external feedback~\citep{pan2023automatically}, this approach does not fully tap into their intrinsic \PR{} capabilities. Moreover, it requires task-specific feedback models or tools, increasing the cost of adapting to a broader range of tasks. Furthermore, current LLMs struggle to self-correct reasoning errors without external feedback~\citep{huang2023large}. Our work aims to unlock the model's inherent Progressive Refinement ability, enabling it to perform \PR{} across all domains without relying on external tools.

\textbf{Prompting for Progressive Refinement}
Various Prompting methods have been introduced to enhance Progressive Refinement, 
such as prompting LLMs to generate explanations and self-correct code~\citep{li2023explainingcompetitivelevelprogrammingsolutions}, 
or encouraging them to generate alternative solutions and revision suggestions~\citep{zhang2024selfcontrastbetterreflectioninconsistent}. 
Some methods iteratively improve outputs by generating feedback through task-specific prompts~\citep{madaan2023selfrefineiterativerefinementselffeedback}, 
or guide models to generate fine-grained feedback in mathematical problem-solving, further enhancing solution accuracy and quality~\citep{xue2023rcot}. 
The Reflexion method enables language models to operate effectively in specific environments by allowing them to reflect and adjust their actions 
when encountering errors~\citep{shinn2023reflexion}. 
However, these approaches often require carefully designed, task-specific prompts or even oracle 
ground-truth answers~\citep{shinn2023reflexionlanguageagentsverbal}, 
making LLMs highly sensitive to evaluating response and achieving optimal performance~\citep{wu2024how}. 
Without external tools, LLMs have limited self-correction capabilities when relying solely on prompting~\citep{huang2023large, zheng2024natural}.

\textbf{Fine-Tuning for Progressive Refinement}
In current \PR{} work, fine-tuning typically relies on reward models or verifiers to assess the accuracy of model outputs based on predefined criteria~\citep{wang2023math, lightman2023let, uesato2022solving}. For instance, some research focuses on improving the model's ability to identify and correct mistakes~\citep{han2024smalllanguagemodelselfcorrect}, while others progressively validate solutions, such as in solving math problems~\citep{uesato2022solvingmathwordproblems}. Additionally, reinforcement learning (RL)~\citep{chen2024self,yuan2024self,rosset2024direct,akyurek-etal-2023-rl4f} has been applied to align model outputs with correct responses. For example, researchers create preference-based datasets to align outputs with human values and reduce harmful content~\citep{wang2024enablinglanguagemodelsimplicitly, rosset2024directnashoptimizationteaching}. Similarly, ROUGE has been used as a reward function in text summarizing tasks to optimize generated summaries~\citep{akyurek-etal-2023-rl4f}.
While these methods effectively train models, they focus on building task-specific datasets and reward functions tailored to particular objectives. In contrast, our approach redefines the fine-tuning objective to bolster the model’s capacity for progressive refinement. Rather than relying on domain-specific datasets, our model is trained to iteratively enhance its responses—starting from initial thoughts and evolving toward increasingly refined answers.

\section{Progressive Thought Refinement Framework}

Our proposed framework, Progressive Thought Refinement (\model{}), comprises two stages, as illustrated in Figure~\ref{fig:main}: (1) Progressive Thought Refinement Dataset Construction and (2) Progressive Weighted Thought-Mask Fine-tuning. The primary objective of this framework is to enhance models' \PR{} abilities, enabling them to handle diverse and unfamiliar tasks without relying on task-specific fine-tuning. Since fine-tuning models for every task is impractical, our approach utilizes general queries, thoughts, and answers to help models comprehend \PR{}. This strategy gradually improves their capacity to tackle complex tasks through progressive refinement.

\subsection{Progressive Thought Refinement Dataset Construction}

In the first stage, we construct a progressive refinement dataset that includes Queries, Thoughts, and Answers. The Thoughts capture a sequence of different reasoning attempts, which may be varied, incomplete, or even incorrect, reflecting the model's initial exploration of the problem. In contrast, the Answers provide more confident and well-reasoned responses. This structured approach helps the model implicitly understand the difference between initial thoughts and improved answers, enabling it to generate more thoughtful and in-depth responses over time.

\subsubsection{Query Preparation}
To enhance the model’s generalization, we avoid creating domain-specific datasets. Instead, we use queries from open-domain general datasets (details in Appendix \ref{appendix:Open-domain}), 
ensuring the model develops general refinement abilities rather than specializing in specific areas. Our data preprocessing involves three key steps. First, we perform data cleaning to remove noise and irrelevant content, such as images or URLs. Second, to prevent data leakage, we exclude domain-specific testing queries during training. Finally, we incorporate traditional SFT data (queries and answers) into our dataset to mitigate the risk of catastrophic forgetting.

\subsubsection{Thought-Answer Preparation}
We strategically select weak and strong models to generate sequences of thoughts and improved answers from an initial query. The objective is to ensure that the final answer is progressively improved through multiple iterations rather than relying on a single-step response. We also employ \textbf{In-Context Learning (ICL)}~\citep{dong2024surveyincontextlearning} and consistency filtering to ensure logical coherence between thoughts and answers.

\forceindent \textbf{Weak-Strong Model Collaborative Selection Criteria}
\label{sec:weak strong}
To ensure the final answer shows significant improvement over the initial thought sequence, we adopt a weak-strong model collaborative selection strategy. Let $\theta_w$ and $\theta_s$ represent the abilities of the weak and strong models, respectively, with the goal of ensuring $\theta_s \gg \theta_w$. We employ three key strategies: \textit{Model Parameter Strength}, \textit{Model Version (New vs.\ Old)}, and \textit{Domain-Specific Fine-Tuning}. These selection strategies ensure the quality of the final answer surpasses that of the previous thoughts. Additionally, we validate that the strong model performs significantly better than the weak model through Wilcoxon significance tests, as shown in Appendix~\ref{appendix:Wilcoxon}.

\forceindent \textbf{Thought Generation by the Weak Model} The weak model generates a sequence of thoughts based on the input query $q_i$, with $\hat{y}_{i,w}^t$ representing the initial thought at the $t$-th attempt. We denote the strong model as $\pi_{\text{strong}, \theta_s}$ and the weak model as $\pi_{\text{weak}, \theta_w}$. These initial thoughts may contain errors but provide a foundation for further refinement:

\vspace{-0.4cm}
\begin{align}
  S_{\text{i, thought}} = \left[\hat{y}_{i,w}^1, \hat{y}_{i,w}^2, \dots, \hat{y}_{i,w}^t\right] = \pi_{\text{weak}, \theta_w}(\cdot \mid q_i).
  \end{align}
  \vspace{-0.4cm}

  Multiple weak models can be used to generate these thoughts, or a single weak model can produce multiple attempts. Since the weak model's thoughts need not be correct, constructing these thoughts remains cost-effective.

\forceindent \textbf{Answer Refinement by the Strong Model} 
To achieve progressive refinement, we leverage the strong model to produce increasingly improved answers. We use ICL to ensure logical coherence between the outputs of the strong and weak models and to avoid randomness. This guides the strong model to generate better answers based on prior thoughts. Specifically, the strong model takes the sequence of thoughts $S_{\text{i, thought}}$ and query $q_i$ as input and generates the final answer $\hat{y}_{i,s,\text{icl}}$:

\vspace{-0.4cm}
\begin{align}
  \hat{y}_{i,s,\text{icl}} = \pi_{\text{strong}, \theta_s}(\cdot \mid S_{\text{i, thought}}, q_i).
\end{align}
\vspace{-0.4cm}

\forceindent\textbf{Thoughts-Answer Consistency Filtering}
To further ensure that the thought process exhibits logical coherence, we apply consistency filtering to remove inconsistent outputs. If the consistency score is below a certain threshold, the pair is considered inconsistent and removed, ensuring that only coherent thought sequences are used for the final output (see Appendix~\ref{appendix:Self-Consistency}).

\subsection{Progressive Weighted Thought-Mask Fine-tuning}

In the second stage, we perform weighted thought-mask fine-tuning using the datasets constructed previously, consisting of the input query $q_i$, the initial thought sequence $S_{\text{i, thought}}$, and the final answer $\hat{y}_{i,s,\text{icl}}$. Formally, the dataset is represented as:

\vspace{-0.4cm}
\begin{align}
  \tilde{\mathcal{D}} = \left\{ \left( q_i, S_{\text{i, thought}}, \hat{y}_{i,s,icl} \right) \right\}_{i=1}^N
\end{align}
\vspace{-0.4cm}

\forceindent \textbf{Thought Mask Mechanism}
\label{sec:thought_mask}
To help the model understand the improvement between the thought process and the answer—rather than focusing solely on the final answer—we introduce a thought mask mechanism. This mechanism selectively hides parts of the thought process during training, as shown in Figure~\ref{fig:main} (B). 
It calculates the loss based only on the accuracy of the refined final answer, ensuring the model focuses on enhancing the quality of its ultimate response. Additionally, we provide refinement instructions (e.g., "Please continue thinking and refine your answer") after each thought process to prompt better refinement in subsequent iterations.

\forceindent \textbf{Weighted Supervised Learning} 
We adopt a weighted supervised learning approach that enables the model to focus on refining its answers by progressively improving its thought process. Specifically, we perform weighted supervised learning that emphasizes both the accuracy of the final answers and the logical consistency of the thought process. The loss function optimizes the model in three key areas: generating accurate final answers, maintaining consistency in reasoning 
and ensuring that the model's confidence increases progressively throughout the thought process.

\vspace{-0.4cm}
\begin{align}
  \label{eq:loss}
    \mathcal{L}_{\text{PTR}}(\theta) &= \sum\limits_{(q_i, S_{i, \text{thought}}, y_n) \in \tilde{\mathcal{D}}} \Bigg[ -\lambda_1 \log \Pr(y_n \mid q_i, S_{i,\text{thought}}; \theta) \nonumber \\
    &\quad + \lambda_2 \sum\limits_{t=2}^{n} \mathcal{F}_{\text{cons}}(y_t, y_{t-1}) + \lambda_3 \sum\limits_{t=1}^{n} \beta_t \left(1 - \Pr(y_t \mid q_i, S_{i,\text{thought}}; \theta)\right) \Bigg].
\end{align}
\vspace{-0.4cm}

Unlike standard supervised fine-tuning, which trains the model to produce a single response $\hat{\mathbf{y}}$ given $\mathbf{x}$, Equation~\ref{eq:loss} focuses exclusively on the accuracy of the final response generated after the thought refinement process. It also ensures that the final response remains logically consistent with the previous thought sequence. We encourage the model to maintain higher confidence in its predictions during subsequent reasoning steps. Here, $\lambda_1$, $\lambda_2$, and $\lambda_3$ are dynamically adjusted according to the model's needs, with their sum constrained to 1.

\section{Experiments}

\label{sec:exps}

The goal of our experiments is to demonstrate the effectiveness of \model{} in enabling language models to progressively enhance their responses. Specifically, we aim to answer the following questions: 
(1) Can the \model{} method activate the model's \PR{} ability? 
(2) Does our method demonstrate generalization? 
(3) Does \PR{} ability emerge during training? 
(4) Is our method robust across different LLMs and instructions? 
(5) How many iterations are required for our method to achieve optimal performance? 

\label{sec:Parameters}

\textbf{Datasets Sources} 
Our model has trained on our \model{} dataset, derived from the WizardLM dataset~\citep{xu2023wizardlm}. After thorough cleaning, we reduced the original dataset from approximately 50k QA pairs to 40k high-quality QA pairs.

\textbf{Evaluation Tasks} 
In our experiments, we perform generalization over ten datasets across different tasks. For general tasks, we use MMLU~\citep{hendrycks2020measuring}, and for coding tasks, we use HumanEval~\citep{chen2021evaluating} (abbreviated as H-Eval). DROP~\citep{dua2019drop} is used for comprehension tasks (abbreviated as Comp), and XSum~\citep{narayan2018dontdetailsjustsummary} is applied for summary tasks. We use GSM8K~\citep{cobbe2021trainingverifierssolvemath} and MATH~\citep{hendrycks2021measuringmathematicalproblemsolving} for math-related tasks. For complex reasoning tasks, we use ARC and GPQA~\citep{rein2023gpqagraduatelevelgoogleproofqa}. For knowledge reasoning, we utilize Winogrande~\citep{sakaguchi2019winograndeadversarialwinogradschema} (abbreviated as Wino) and CommonsenseQA~\citep{talmor2019commonsenseqaquestionansweringchallenge} (abbreviated as Comm).

\textbf{Evaluation Settings} 
We use greedy decoding (with temperature set to 0) for final generation, as lower temperature yields better performance shown in Appendix~\ref{appendix:temp_stop}. We utilize zero-shot prompting~\citep{kojima2023largelanguagemodelszeroshot} for both answer sampling and evaluations, observing that zero-shot prompting outperforms few-shot prompting for LLMs fine-tuned on specific tasks.
All of our experiments are conducted on workstations equipped with eight NVIDIA A800 PCIe GPUs with 80GB memory, running Ubuntu 20.04.6 LTS and PyTorch 2.0.1.

\textbf{Baselines}
We compare our model with base models and prior approaches:
(1) \textbf{Prompt}: Directly prompting the model to refine its answer~\citep{huang2023large}. 
(2) \textbf{IFT}: Instruction Fine-Tuning by directly fine-tuning the input-output pairs from strong models on the PRD dataset to show that improvements are not due to knowledge distillation. 
(3) \textbf{RL}: Perform one reinforcement learning training~\citep{wu2024progressregressselfimprovementreversal} iteration on the PRD dataset to compare with our method. Specifically, we use the thoughts and answers of the PRD dataset to construct preference data, and prefer model to produce stronger answers through DPO~\citep{rafailov2024directpreferenceoptimizationlanguage}.
We compare these methods on the PRD dataset under the same settings as in the previous section. Detailed settings are in Appendix~\ref{appendix:setting}.

\setlength\tabcolsep{1pt}
\begin{table*}[t]
\small
  \centering
    \begin{tabular}{llgvgvgvggvvg}
      \toprule\toprule
    \multirow{3}{*}{\textbf{Method}} & 
    \multirow{3}{*}{\parbox{0.8cm}{\textbf{Iters}}} & 
    \multicolumn{1}{c}{\textbf{Gene.}} & 
    \multicolumn{1}{c}{\textbf{Code}} & 
    \multicolumn{1}{c}{\textbf{Comp.}} &
    \multicolumn{1}{c}{\textbf{Sum.}} &
    \multicolumn{2}{c}{\textbf{Math}} &
    \multicolumn{2}{c}{\textbf{Reasoning}} &
    \multicolumn{2}{c}{\textbf{Knowledge}} &
    \multicolumn{1}{c}{\textbf{ }} 
     \\
    \cmidrule(lr){3-3} 
    \cmidrule(lr){4-4}
    \cmidrule(lr){5-5}
    \cmidrule(lr){6-6}
    \cmidrule(lr){7-8}
    \cmidrule(lr){9-10} 
    \cmidrule(lr){11-12}
    \cmidrule(lr){12-12}
    & & 
    \multicolumn{1}{c}{\textbf{MMLU}} & 
    \multicolumn{1}{c}{\textbf{H-Eval}} & 
    \multicolumn{1}{c}{\textbf{DROP}} &
    \multicolumn{1}{c}{\textbf{Xsum}} &
    \multicolumn{1}{c}{\textbf{GSM8k}} &
    \multicolumn{1}{c}{\textbf{Math}} &
    \multicolumn{1}{c}{\textbf{ARC}} &
    \multicolumn{1}{c}{\textbf{GPQA}} &
    \multicolumn{1}{c}{\textbf{Wino}} &
    \multicolumn{1}{c}{\textbf{Comm}} &
    \multicolumn{1}{c}{\textbf{Avg}}\\
    \cmidrule(lr){3-3}
    \cmidrule(lr){4-4}
    \cmidrule(lr){5-5}
    \cmidrule(lr){6-6}
    \cmidrule(lr){7-7}
    \cmidrule(lr){8-8}
    \cmidrule(lr){9-9}
    \cmidrule(lr){10-10}
    \cmidrule(lr){11-11}
    \cmidrule(lr){12-12}
    \cmidrule(lr){12-12}
    & 
    & 
    \multicolumn{1}{c}{\textbf{Acc}} & 
    \multicolumn{1}{c}{\textbf{Pass@1}} &
    \multicolumn{1}{c}{\textbf{Acc}} &
    \multicolumn{1}{c}{\textbf{Sim}} &
    \multicolumn{1}{c}{\textbf{Acc}} &
    \multicolumn{1}{c}{\textbf{Acc}} &
    \multicolumn{1}{c}{\textbf{Acc}} &
    \multicolumn{1}{c}{\textbf{Acc}} &
    \multicolumn{1}{c}{\textbf{Acc}} &
    \multicolumn{1}{c}{\textbf{Acc}} &
    \multicolumn{1}{c}{} \\
    \midrule
    
    \rowcolor[rgb]{ .949,  .953,  .961} \multicolumn{13}{c}{\textit{Qwen2-7b}} \\

    \multirow{2}{*}{Prompt} 
    & Iter.1& 
    57.1          & 56.1          & 20.9          & 47.3          & 79.1            & 48.2          & 60.6          & 24.6            & 66.8          & 55.7      & 51.6\\
    & Iter.2 &                                                                            
      50.1          & 37.6          & 18.7          & 43.2          & 78.4            & 47.6          & 37.9          & 22.3            & 50.4          & 42.1      & 42.8\\
    \midrule          
    \multirow{2}{*}{IFT} 
     & Iter.1& 
    57.7          & 50.2          & 21.1          & 45.5          & 75.4            & 45.6          & 54.9          & 22.3            & 66.8          & 46.1      & 48.5\\
    & Iter.2 &  
    52.4          & 40.2          & 17.2          & 37.9          & 71.0            & 43.2          & 36.6          & 21.9            & 62.8          & 40.3      & 42.3\\
    \midrule 
    \multirow{2}{*}{RL} 
    & Iter.1& 
   56.5          & 48.3          & 21.7          & 47.6          & 71.2            & 47.3          & 60.4           & 20.3            & 65.0          & 51.6      & 48.9\\
   & Iter.2 &  
   55.1          & 42.2          & 20.9          & 44.3          & 58.6            & 44.5          & 35.3          & 20.9            & 63.8          & 43.5      & 42.9\\
   \midrule 
    \multirow{2}{*}{\textbf{\model}(our)} 
     & Iter.1 & 
    59.2          & 52.3          & 19.0          & 45.9          & 76.7            & 47.6          & 58.6          &23.2             & 66.4   & 47.9    & 49.6\\
    & Iter.2 &
    \textbf{64.1} & 57.2          & 21.2          & \textbf{49.8} & 79.6            & 48.6          & 62.7          & \textbf{25.6}   & 66.4            & 54.9    & 53.0\\
    & Iter.3 &
    63.2          & \textbf{57.6} & \textbf{21.5} & 49.6          & \textbf{79.9}   & \textbf{48.9} & \textbf{65.2} & \textbf{25.6}   & \textbf{66.8}            & \textbf{56.5}  & \textbf{53.5} \\
        \midrule

    \rowcolor[rgb]{ .949,  .953,  .961} \multicolumn{13}{c}{ \textit{Llama3-8B}} \\
    \multirow{2}{*}{Prompt} 
    & Iter.1 & 
    66.4 & 54.0 & 47.3 & 64.5 & 76.4 & 25.1 & 75.1 & 34.6 & 67.9 & 41.6 & 55.2 \\
    & Iter.2 &
    34.4 & 50.1 & 35.9 & 62.1 & 70.5 & 20.9 & 56.4 & 30.1 & 66.8 & 43.9 & 47.1 \\
    \midrule 
    \multirow{2}{*}{IFT} 
     & Iter.1 & 
    49.1 & 38.4 & 52.8 & 47.9 & 55.4 & 21.3 & 63.0 & 34.1 & 63.3 & 37.1  & 46.2\\
    & Iter.2 &
    40.4 & 34.2 & 24.7 & 42.9 & 51.1 & 18.6 & 54.4 & 28.8 & 62.2 & 28.7 & 38.5 \\
    \midrule 
    \multirow{2}{*}{RL} 
    & Iter.1 & 
    51.8 & 32.4 & 46.8 & 65.9 & 61.3 & 22.3 & 67.6 & 33.7 & 62.0 & 60.1  & 50.3\\
   & Iter.2 &
   39.9 & 30.2 & 40.7 & 40.9 & 57.3 & 19.3 & 55.7 & 30.6 & 53.8 & 55.8 & 42.4 \\
   \midrule 
    \multirow{2}{*}{\textbf{\model}(our)} 
     & Iter.1 & 
    59.6 & 54.0 & 49.0 & 62.4 & 76.4 & 21.3 & 73.0 & 34.1 & 68.6 & 60.0 & 55.8 \\
    & Iter.2 &
     68.4         & 55.2          & \textbf{49.0} &         65.7  & 79.2          & 24.7          & \textbf{77.1} & \textbf{36.2} & \textbf{70.1} & 60.5  & 58.6\\
    & Iter.3 &
    \textbf{68.6} & \textbf{55.4} & 48.6          & \textbf{66.1} & \textbf{79.6} & 24.9 & 77.0          & 36.1          & 67.9          & \textbf{61.3} & \textbf{58.6} \\

    \bottomrule \bottomrule
    \end{tabular}
  \caption{
    Main experimental results about our approach and other baselines 
    across various domains. 
    We experiments on two difference structures LLMs( Qwen2-7b and Llama3-8B ).
    We also run 2 iteration on different baselines and 3 iteration on our approach.
    Itertaion 1 means the first answer to the question, and we construct the format of refining instruction with the previous answer which is introduced in Method~\ref{sec:thought_mask}.
    We denoted Acc: accuracy. Pass@1:testing on code. Sim: similarity similared calculated by BGE-m3.
    These results suggest that our \model{} excels at performing well across multiple attempts. 
    By trading off some accuracy on the first attempt, 
    it significantly enhances the model's ability to improve in subsequent iterations.}
  \label{tab:main_results}
  \vspace{-0.7cm}
\end{table*}

\subsection{Can the \model{} method activate the model's \PR{} ability?}
\label{sec:Activating}

\textbf{\model{} Activates Progressive Refinement Ability} 
As shown in Table~\ref{tab:main_results}, to emphasize the \PR{} ability, 
we conduct tests on a broad range of tasks.
The result demonstrates that our \model{} activate
models substantially 
refine their responses across multiple iterations in the majority of tasks.
For instance, in the Qwen2-7B model, accuracy on MMLU increased by
7.0\%, from 57.1\% (Base model Prompt Iteration 1) to 64.1\% (\model{} Iteration 3). 
On several additional tasks, \model{} also showed improvements, 
with the average score across all tasks increasing by 3.9\%-rising from 49.6\% to 53.5\%.
However, the \textbf{Prompting} method results show that both two base models degrade 
in performance when asked to refine, producing worse answers compared to initial responses. 
These results indicate that \model{} effectively enables 
base models to improve based on previous thoughts.

\textbf{\model{} vs. Knowledge Distillation (IFT)} 
We also compare our \model{} with Knowledge Distillation (IFT).
We find that \model{} is not equivalent to knowledge distillation. 
At the first iteration, we observe that when models are trained on general datasets rather 
than domain-specific tasks, 
its initial performance tends to decline at first. 
We found that this performance drop largely stems 
from supervised fine-tuning amplifying the initial biases of the base model. 
When trained on general datasets, the base models tend to accumulate biases 
that may not apply to specific domains, 
leading to poorer performance on domain-specific tasks.
For example, in ARC tasks, accuracy drops from 60.6\% (Base model Iteration 1) to 54.9\% (IFT Iteration 1). This demonstrates that our datasets are clean and that knowledge distillation alone cannot improve accuracy in specific domains.
Instead of correcting these biases at first response, 
we focus on correcting them through iterative refinement.
However, the IFT approach fails to activate the model’s \PR{} ability 
and does not significantly increase the performance after the first attempts.
On some reasoning benchmarks, such as CommonsenceQA, 
The IFT approach does not perform a better response at the second iteration (40.3\%) compared to their first attempt (46.1\%).
In contrast, \model{} approach improves through iterative attempts 
without an approach on domain-specific knowledge.
This suggests that our method is not simply \textbf{distilling knowledge} but effectively \textbf{activating} the model to refine outputs and enhance performance through self-driven iterative improvement.

\textbf{Refinement beyond Correction} 
Deeper analysis reveals that in open-ended tasks without clear ground truth, 
LLMs refine responses to be more thoughtful and comprehensive, regardless of correctness. 
For example, in the code task shown in Figure~\ref{fig:code_example}, the LLM iteratively improves its response over three iterations, considering additional aspects of the problem. This highlights \model{}'s ability to enhance not just correctness but also the quality and usability of outputs (Shown in Appendix~\ref{app:case-study}).

\lstset{
    language=Python,             
    basicstyle=\ttfamily,   
    keywordstyle=\bfseries\color{Pink},
    commentstyle=\itshape\color{Bluedeep},
    stringstyle=\color{Orgdeep},
    frame=single,           
    breaklines=true,        
    xleftmargin=0pt         
}
\vspace{-2mm}

\begin{figure}[t]
   \center
   \begin{tcolorbox}[fontupper=\scriptsize, width=1\linewidth, colback=white!90, opacityback=0.96] 
      \vspace{-0.25cm}
      
      \begin{tikzpicture}[remember picture,overlay]
         
      \draw[->, line width=0.5mm, Orgdeep] (-0.3,-1) -- (-0.3,-2);
      \draw[->, line width=0.5mm, Orgdeep] (-0.3,-2.5) -- (-0.3,-3.5);

      \end{tikzpicture}

      \begin{lstlisting}
def calculation_average(numbers):   # iteration 1
   total = sum(numbers) 
   return total / len(numbers) # returning the output
         \end{lstlisting}
         \vspace{-0.25cm}
\begin{lstlisting}
def calculation_average(numbers):  # iteration 2
   total = sum(numbers) 
   return total/len(numbers) if len(numbers) > 0 else None # Adds empty lists check
\end{lstlisting}
\vspace{-0.25cm}
\begin{lstlisting}
def calculation_average(numbers):  # iteration 3 
   if not isinstance(numbers, list):
      raise TypeError("Input_must_be_a_list") # Introduces type checking
   if len(numbers) == 0:   # Raises specific errors with informative messages
      raise ValueError("List_cannot_be_empty") # Adds protection against empty lists
   return sum(numbers)/len(numbers)
\end{lstlisting}   
\vspace{-0.4cm}
   \end{tcolorbox}  
   \caption{Code example shows \model{} can refine beyond correction.
   The \model{} goes through three rounds, providing higher quality response for each iterations. In first interation, model return with simply output.
   In second interation, model add more details like considering the empty list. 
   In third interation, model structured the code and futher add type checking and errors information.} 
   \label{fig:code_example}
  
\end{figure}

  \begin{table*}[h]
    \small
      \centering
        \begin{tabular}{ll|cccccccccc}
          \toprule\toprule
        \multirow{3}{*}{\textbf{Prompt}} & 
        \multirow{3}{*}{\parbox{0.8cm}{\textbf{Iters}}} & 
        \multicolumn{1}{c}{\textbf{General}} & 
        \multicolumn{1}{c}{\textbf{Code}} & 
        \multicolumn{1}{c}{\textbf{Compreh.}} &
        \multicolumn{1}{c}{\textbf{Summary}} &
        \multicolumn{2}{c}{\textbf{Math}} &
        \multicolumn{2}{c}{\textbf{Reasoning}} &
        \multicolumn{2}{c}{\textbf{Knowledge}} \\
        \cmidrule(lr){3-3} 
        \cmidrule(lr){4-4}
        \cmidrule(lr){5-5}
        \cmidrule(lr){6-6}
        \cmidrule(lr){7-8}
        \cmidrule(lr){9-10} 
        \cmidrule(lr){11-12}
        & & 
        \multicolumn{1}{c}{\textbf{MMLU}} & 
        \multicolumn{1}{c}{\textbf{H-Eval}} & 
        \multicolumn{1}{c}{\textbf{DROP}} &
        \multicolumn{1}{c}{\textbf{Xsum}} &
        \multicolumn{1}{c}{\textbf{GSM8k}} &
        \multicolumn{1}{c}{\textbf{Math}} &
        \multicolumn{1}{c}{\textbf{ARC}} &
        \multicolumn{1}{c}{\textbf{GPQA}} &
        \multicolumn{1}{c}{\textbf{Wino*}} &
        \multicolumn{1}{c}{\textbf{Comm*QA}} \\
        \cmidrule(lr){3-3}
        \cmidrule(lr){4-4}
        \cmidrule(lr){5-5}
        \cmidrule(lr){6-6}
        \cmidrule(lr){7-7}
        \cmidrule(lr){8-8}
        \cmidrule(lr){9-9}
        \cmidrule(lr){10-10}
        \cmidrule(lr){11-11}
        \cmidrule(lr){12-12}
        & 
        & 
        \multicolumn{1}{c}{\textbf{Acc}} & 
        \multicolumn{1}{c}{\textbf{Pass@1}} &
        \multicolumn{1}{c}{\textbf{Acc}} &
        \multicolumn{1}{c}{\textbf{Rank}} &
        \multicolumn{1}{c}{\textbf{Acc}} &
        \multicolumn{1}{c}{\textbf{Acc}} &
        \multicolumn{1}{c}{\textbf{Acc}} &
        \multicolumn{1}{c}{\textbf{Acc}} &
        \multicolumn{1}{c}{\textbf{Acc}} &
        \multicolumn{1}{c}{\textbf{Acc}} \\
        \midrule

    
        \multirow{4}{*}{\textbf{Prompt1}} ~
         & Iter.1 & 
        58.7          & 53.9         & 19.2          & 46.9          & 75.1            & 47.7          & 50.5          & 24.6             & 66.7   & 46.7  \\
      
        & Iter.2 &
        \textcolor{antiquebrass}{63.4} & \textcolor{antiquebrass}{57.6}         & \textcolor{antiquebrass}{20.7}          & \textcolor{gray}{46.6} & \textcolor{antiquebrass}{77.7}            & \textcolor{antiquebrass}{48.8}          & \textcolor{antiquebrass}{60.6}  & \textcolor{antiquebrass}{25.4}            & \textcolor{gray}{66.1}            & \textcolor{antiquebrass}{51.5}  \\
        & Iter.3 &
        \textcolor{antiquebrass}{63.2}          & \textcolor{antiquebrass}{57.6} & \textcolor{antiquebrass}{20.9} & \textcolor{antiquebrass}{49.8}          & \textcolor{antiquebrass}{79.2}   & \textcolor{antiquebrass}{50.2} & \textcolor{antiquebrass}{61.3}          & \textcolor{antiquebrass}{24.9}            & \textcolor{gray}{65.9}            & \textcolor{antiquebrass}{54.7}  \\
        & Iter.4 &
        \textcolor{antiquebrass}{63.3}          & \textcolor{antiquebrass}{57.6}          & \textcolor{antiquebrass}{21.7} & \textcolor{antiquebrass}{49.9}          & \textcolor{antiquebrass}{78.1}            & \textcolor{antiquebrass}{50.6}          & \textcolor{antiquebrass}{62.8}  & \textcolor{antiquebrass}{25.6}            & \textcolor{gray}{66.6}            & \textcolor{antiquebrass}{55.7}  \\
       \midrule

       \multirow{5}{*}{\textbf{Prompt2}} 
       & Iter.1 & 
       58.7          & 53.0         & 19.2          & 46.6          & 75.1            & 47.0          & 44.5          & 25.4            & 66.7   & 46.7  \\
     
       & Iter.2 &
       \textcolor{antiquebrass}{63.4} & \textcolor{gray}{52.8}         & \textcolor{antiquebrass}{22.1}          & \textcolor{antiquebrass}{48.9}                   & \textcolor{antiquebrass}{77.5}            & \textcolor{antiquebrass}{47.2}  & \textcolor{antiquebrass}{61.1}          & \textcolor{gray}{25.4}           & \textcolor{antiquebrass}{68.8}            & \textcolor{antiquebrass}{52.1}  \\
       & Iter.3 &
       \textcolor{antiquebrass}{62.7}          & \textcolor{antiquebrass}{57.9}         & \textcolor{antiquebrass}{22.5} & \textcolor{antiquebrass}{49.8}          & \textcolor{antiquebrass}{76.8}            & \textcolor{antiquebrass}{47.9}          & \textcolor{antiquebrass}{59.1} & \textcolor{antiquebrass}{25.6}   & \textcolor{antiquebrass}{68.2}   & \textcolor{antiquebrass}{50.4}  \\
       & Iter.4 &
       \textcolor{antiquebrass}{62.8}  & \textcolor{antiquebrass}{57.6}          & \textcolor{antiquebrass}{22.4}          & \textcolor{antiquebrass}{49.6} & \textcolor{antiquebrass}{77.5}            & \textcolor{antiquebrass}{47.8}          & \textcolor{antiquebrass}{60.0}          & \textcolor{antiquebrass}{25.8}   & \textcolor{antiquebrass}{67.5}            & \textcolor{antiquebrass}{53.1}  \\
       
       \midrule

       \multirow{5}{*}{\textbf{Prompt3}} 
      & Iter.1 & 
       58.7          & 52.3         & 19.2            & 48.8          & 75.1            & 47.5          & 47.5          & 23.6             & 66.7   & 46.8  \\
      
       & Iter.2 &
       \textcolor{antiquebrass}{63.4} & \textcolor{antiquebrass}{57.6}         & \textcolor{antiquebrass}{21.6}            & \textcolor{gray}{48.2}          & \textcolor{antiquebrass}{78.3}   & \textcolor{antiquebrass}{48.6}          & \textcolor{antiquebrass}{59.4}          & \textcolor{antiquebrass}{25.6}   & \textcolor{gray}{66.4}            & \textcolor{antiquebrass}{50.2}  \\
       & Iter.3 &
       \textcolor{antiquebrass}{62.9}          & \textcolor{antiquebrass}{57.9} &        \textcolor{antiquebrass}{21.9}    & \textcolor{antiquebrass}{49.6} & \textcolor{antiquebrass}{78.1}            & \textcolor{antiquebrass}{48.7}          & \textcolor{antiquebrass}{62.1}          & \textcolor{antiquebrass}{25.0}   & \textcolor{antiquebrass}{67.0}            & \textcolor{antiquebrass}{55.0}  \\
       & Iter.4 &
       \textcolor{antiquebrass}{63.3}          & \textcolor{antiquebrass}{57.8}          & \textcolor{antiquebrass}{22.3}  & \textcolor{antiquebrass}{49.6} & \textcolor{antiquebrass}{77.9}            & \textcolor{antiquebrass}{49.6} & \textcolor{antiquebrass}{63.2}  & \textcolor{antiquebrass}{25.4}            & \textcolor{antiquebrass}{67.2}            & \textcolor{antiquebrass}{53.8}  \\
       \bottomrule
    
        \bottomrule
        \end{tabular}
      \caption{Results of \model{} with different prompts: (1) Assume that this thought could be either correct or incorrect. Carefully review the thought and provide a better answer.
      (2) Review your previous thought and assess whether it's correct. Then, provide a better response based on your answer.
      (3) Regardless of whether your previous thought is correct or not, provide a better answer.
      Iteration 1 represent the initial answer to the question.
      Iteration 2 to 4 represent the model's improvement over the initial answer.
      Notibly, the model is not trained with these prompts.}
      \label{tab:prompt_results}
      \vspace{-0.4cm}
    \end{table*}

\subsection{Does our method demonstrate generalization?}
\label{sec:generalization}

\textbf{\model{}  vs. Other Progressive Refinement methods} 
Unlike previous approaches, our method activates the model's inherent \PR{} ability rather than merely boosting accuracy in 
specific domains. To validate \model{}'s generalization capability, we use datasets with general queries and evaluate whether the model 
can iteratively refine responses across various tasks. 
As seen in Table~\ref{tab:main_results}, our model refines responses across multiple iterations, significantly improving accuracy across tasks, and demonstrating effective generalization.
We also compare \model{} with other \PR{} methods like RL to assess generalization. Our results show that methods like RL, 
when fine-tuned only on general-domain tasks, fails to activate iterative refinement in specialized tasks, often showing decreased accuracy. 
This suggests that our method is more robust in diverse environments, 
as it enables the model to iteratively refine its responses without being limited to domain-specific fine-tuning. 
By leveraging the model's inherent progressive refinement capabilities, 
\model{} achieves consistent improvements across a wide range of tasks.

\subsection{Is our method robust across different LLMs and instructions?}

\textbf{Prompt Robustness}
We also evaluated \model{} robustness with different prompts and LLMs.
Table~\ref{tab:prompt_results} shows the model's performance using three different prompts across various tasks, 
refined over four iterations. 
Across all prompts, 
we find that \model{} achieves iterative improvement across different prompts.
Specifically, 
In the math (GSM8K) tasks, \model{} is well-performed(78.1\%) compared with initial responses (75.1\%).
On reasoning tasks (ARC), \model{} see substantial improvements, especially with Prompts 1 (62.8\%) and Prompts 3 (63.2\%). 
DROP tasks also improve steadily, with accuracy increasing to 22.5\% by Iteration 3 in Prompt 2.
Our approach enables the model to learn from previous thoughts, 
rather than relying on the instruction used during training. 
This \model{} enables the model to consistently improve its performance on different prompts, 
demonstrating the robustness of the \model{} mechanism.

\textbf{LLMs Robustness} 
The table~\ref{tab:main_results} also demonstrates that both Llama3-8B and Qwen2-7B exhibit robustness across different prompts and tasks. While Llama3-8B often outperforms Qwen2-7B, both models show consistent improvements with iterative refinement. This robustness ensures that \model{} can be applied effectively to a wide variety of open-source LLMs.

\vspace{2mm}
\begin{figure}[t]
   \centering
   \includegraphics[width=1\textwidth]{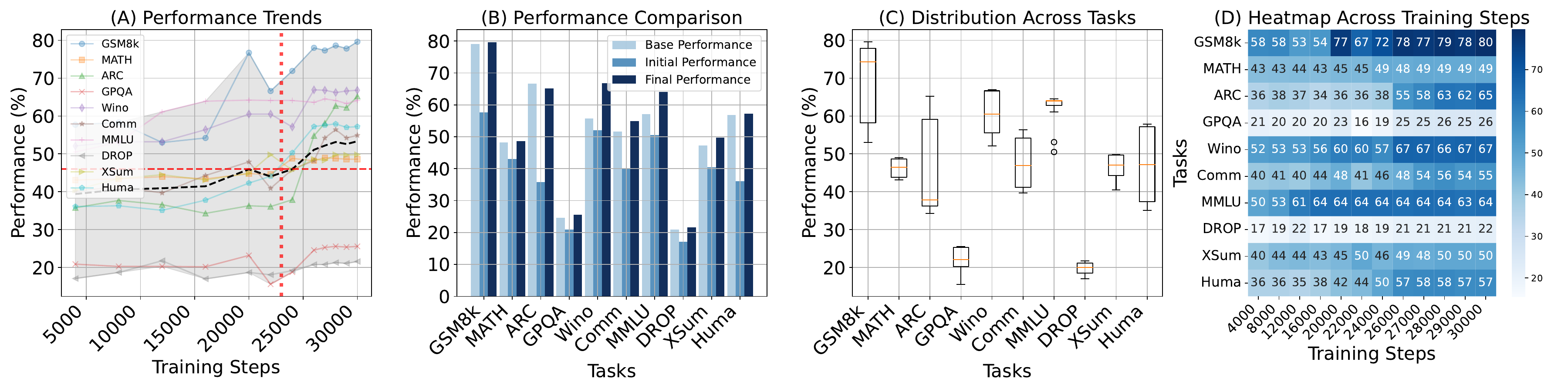}
   \caption{Plot A: Multi-line plot showing the performance trends for multiple tasks, along with average performance and variance.
   Plot B: Bar plot comparing initial, base, and final performance for each task.
   Plot C: Box plot displaying the performance distribution across tasks.
   Plot D: Heat map representing task performance across training steps. }
   \label{fig:Training_Steps}
\end{figure}
\vspace{-0.4cm}
\vspace{2mm}

\subsection{Does \PR{} ability exhibit emergence during training?}
\textbf{Overall Performance} 
Figures (A) and (B) show a clear upward trend in performance, as shown in Figure ~\ref{fig:Training_Steps} shown. Notably, after 24,000 training steps (equivalent to 93 million tokens), significant improvements indicate the emergence of inference capabilities. As training continues, we observe that the average performance of \model{} increases from 40.1\% to 55.6\%, showing an overall improvement across different tasks.

\textbf{Task Complexity and Learning Curve}
We also find that tasks of varying difficulty exhibit different emergence timings and improvement rates. 
Plots (C) and (D) reveal that simpler tasks such as MMLU and DROP show early and steady improvements around 22,000 steps.
More complex inference tasks such as ARC and GPQA exhibit delayed emergence, 
with ARC improving from 36.3\% to 65.2\% and GPQA from 23.2\% to 25.6\% after 24,000 steps.
This shows that as training continues, the model's ability to handle complex reasoning and other tasks significantly improves, 
showing clear emergent behavior in different task types.

\subsection{How many thinking steps are required to achieve optimal performance?}

\begin{figure}[t]
   \centering
   \includegraphics[width=0.9\textwidth]{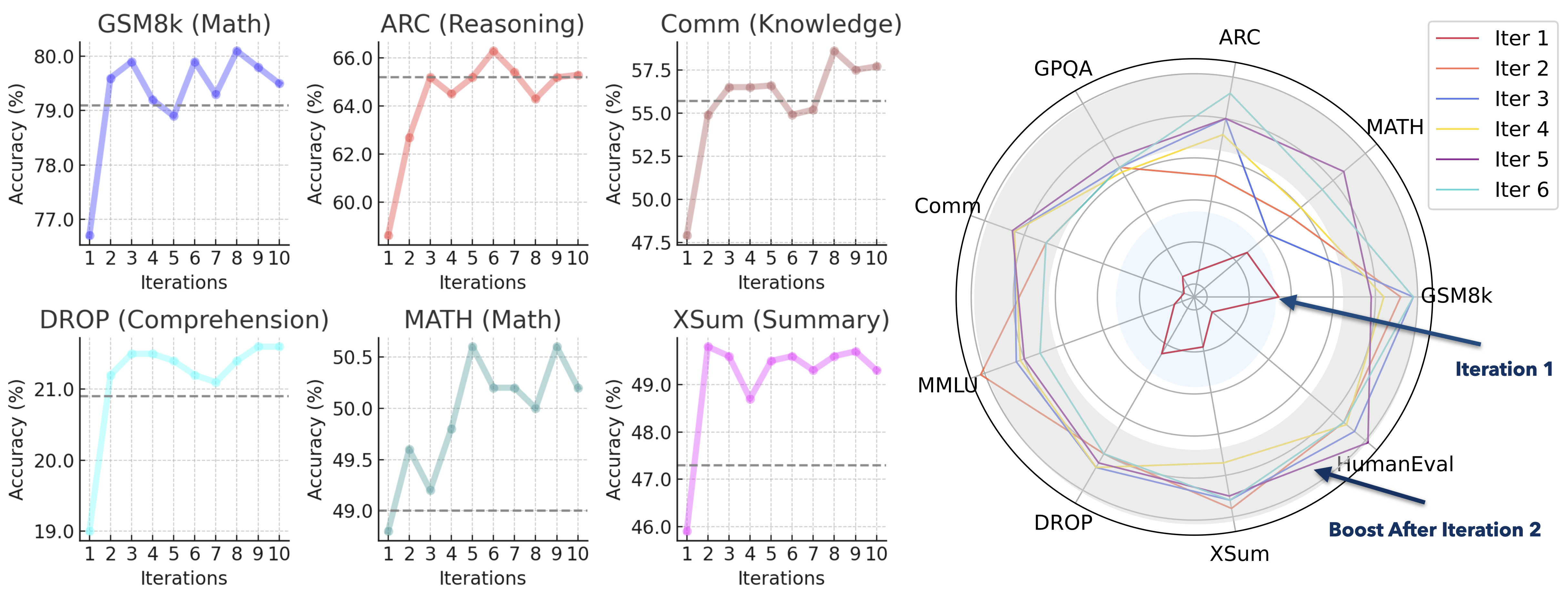}
   \caption{
      Performance of \model{} over ten iterations across different tasks. 
      Performance of \model{} over ten iterations across different tasks. The left plots show accuracy improvements in mathematical reasoning (GSM8K and MATH), reasoning tasks (ARC, GPQA, Winogrande, CommonsenseQA), comprehension tasks (MMLU, DROP, XSum), and coding tasks (HumanEval). More details are in Appendix~\ref{appendix:Iteration-study}. 
      Baseline performance is indicated by dashed lines. The right plots show performance over six iterations with radar charts, illustrating improvement over the first three iterations. 
         }
\label{fig:iterations_test}

\end{figure}

We investigate how iterative thinking steps influence performance across tasks by conducting experiments over ten iterations using the Qwen2-8B model. 
Figure~\ref{fig:iterations_test} illustrates performance trends.

\textbf{Improvements in the First Three Iterations} 
In the first three iterations, we saw significant improvements in model performance. In the mathematical reasoning task \textbf{GSM8K}, the accuracy improved from 75.0\% in the first iteration to 79.9\% in the second iteration. Similarly, the \textbf{ARC} dataset improves from 58.6\% to 65.2\% in the third iteration. This shows that \model{} quickly refines its problem-solving through progressive refinement.

After the third iteration, the performance improvements for most tasks stabilize. In \textbf{GSM8K}, the accuracy fluctuates slightly between the third and tenth iterations, ranging from 79.9\% to 80.1\%. In \textbf{MATH}, the accuracy remains around 50.2\% to 50.6\% after reaching a peak in the second iteration. This indicates that the marginal gains decrease over time, indicating that the performance ceiling of the model is converging.

\textbf{Sustained Performance Without Overfitting} 
\model{} consistently improves across iterations without signs of overfitting. Performance remains stable or improves slightly, with no notable declines. For instance, in \textbf{DROP} and \textbf{XSum}, accuracy increases from 19.0\% and 45.9\% to 21.6\% and 49.7\%, respectively, over ten iterations.

\textbf{More Computation for Hard Tasks} 
Complex tasks benefit more from iterative thinking and may require additional iterations for optimal performance. Accuracy in \textbf{CommonsenseQA} improves from 47.9\% to 58.6\% by the eighth iteration, suggesting that tasks with higher cognitive demands allow \model{} to leverage iterative refinement more effectively. While \textbf{GSM8K} reaches near-optimal performance within a few iterations, tasks like \textbf{MATH} require more computation to achieve substantial gains, likely due to the challenging nature of logical reasoning involved.

\section{Conclusion}
We propose \model{}, an approach designed to stimulate the progressive thought refinement capabilities inherent in LLMs,
allowing them to improve their responses through multiple rounds of iterations.
\model{} adopts an annotation-free strategy to gradually build refined thoughts and answers through a weak and strong models collaborative selection process, 
and combines thought-answer consistency filtering to ensure logical coherence.
Our weighted thought mask fine-tuning further activates the model's internal refinement ability by learning the improvement from initial thoughts to refined answers.
Experimental results show that \model{} simply trained with general open-domain datasets, but significantly improves the model's \PR{} capabilities in ten different tasks, including knowledge reasoning, code generation, and mathematical reasoning, achieving a generalization level not observed by previous methods.
\newpage

\bibliography{iclr2025_conference}
\bibliographystyle{iclr2025_conference}

\newpage
\appendix
\label{sec:appendix}

\section{Method}
\label{appendix:method}

\subsection{Self-Consistency Filtering}
\label{appendix:Self-Consistency}

In each iteration of thought generation, we apply multiple sampling techniques to generate several candidate thoughts. These candidate thoughts undergo a consistency check against the final answer to ensure logical coherence throughout the thought process.

\subsubsection{n-Sampling for Thought Generation}

For each query $q_i$, we perform \textbf{n-sampling} to generate \( N \) candidate thoughts at each step of the thought generation process. These thoughts denoted as \( \hat{y}_{i,w}^t \), represent the $m$-th sample at the $t$-th attempt, and they collectively form the set of potential thought sequences.

\subsubsection{Consistency Filtering}

To evaluate the consistency between the thought sequences and the final answer, we vectorize the thoughts and the answer using two complementary techniques:
\begin{itemize}
    \item \textbf{Distinct N-grams}: This method captures surface-level similarities by extracting unique N-grams from the thought sequences and the final answer. Higher distinct N-gram overlap between two texts suggests greater structural similarity.
    \item \textbf{Sentence-BERT embeddings}: This technique captures semantic similarities by embedding both the thought sequences and the final answer into a shared vector space using Sentence-BERT~\citep{reimers2019sentencebertsentenceembeddingsusing}. This enables a deeper comparison of the meaning conveyed by the thoughts and the answer.
\end{itemize}

The similarity between each thought \( y_j \) and the final answer \( \hat{y}_i \) is computed as a weighted combination of the N-gram similarity and the Sentence-BERT embedding similarity:

To refine the thought sequences, we apply a consistency score between each generated thought and the final answer using the \textbf{Consistency Function} \( \mathcal{F}_{\text{cons}}(y_j, \hat{y}_i) \), which measures the similarity between each thought and the final answer:

\begin{align}
\mathcal{F}_{\text{cons}}(y_j, \hat{y}_i) = \alpha_1 \cdot \text{N-gram}(y_j, \hat{y}_i) + \alpha_2 \cdot \text{BERT}(y_j, \hat{y}_i)
\end{align}
Here, \( \alpha_1 \) and \( \alpha_2 \) are weighting factors that balance the importance of N-gram similarity and sentence-BERT embeddings.
This function calculates the overall similarity between a thought \( y_j \) and the final answer \( \hat{y}_i \), with a higher value indicating better consistency.

Formally, the consistency score for the entire thought sequence \( S_{\text{thought}}^t = \{\hat{y}_{i,w}^1, \hat{y}_{i,w}^2, \dots, \hat{y}_{i,w}^t\} \) with respect to the final answer \( \hat{y}_i \) is computed as:

\begin{align}
C(S_{\text{thought}}^t, \hat{y}_i) = \frac{1}{|S_{\text{thought}}^t|} \sum_{j=1}^{t} \mathbb{I}\left( \mathcal{F}_{\text{cons}}(y_j, \hat{y}_i) \geq \delta \right)
\end{align}

Here, \( \mathbb{I}(\cdot) \) is an indicator function that returns 1 if the consistency score \( \mathcal{F}_{\text{cons}}(y_j, \hat{y}_i) \geq \delta \), and 0 otherwise. The overall score \( C(S_{\text{thought}}^t, \hat{y}_i) \) represents the average consistency between the thought sequence and the final answer.

\subsection{Temperature Adjustment}
\label{appendix:temp_stop}
This graph illustrates the performance of a model under different temperature settings during inference, measured over three iterations in terms of accuracy. The comparison includes four scenarios: fixed temperature at 0, the fixed temperature at 0.4, fixed temperature at 0.8, and gradually decaying temperature. The main findings are as follows:
The graph clearly shows that setting the temperature to 0 yields the best performance. A temperature of 0 ensures that the model generates deterministic outputs at every step, leading to more reliable and stable results. Higher temperatures (such as 0.4 and 0.8) introduce randomness into the process, reducing overall accuracy. The decaying temperature approach improves accuracy over time but does not surpass the performance of a fixed temperature of 0.

\begin{figure}[t]
  \centering
  \includegraphics[width=0.7\linewidth]{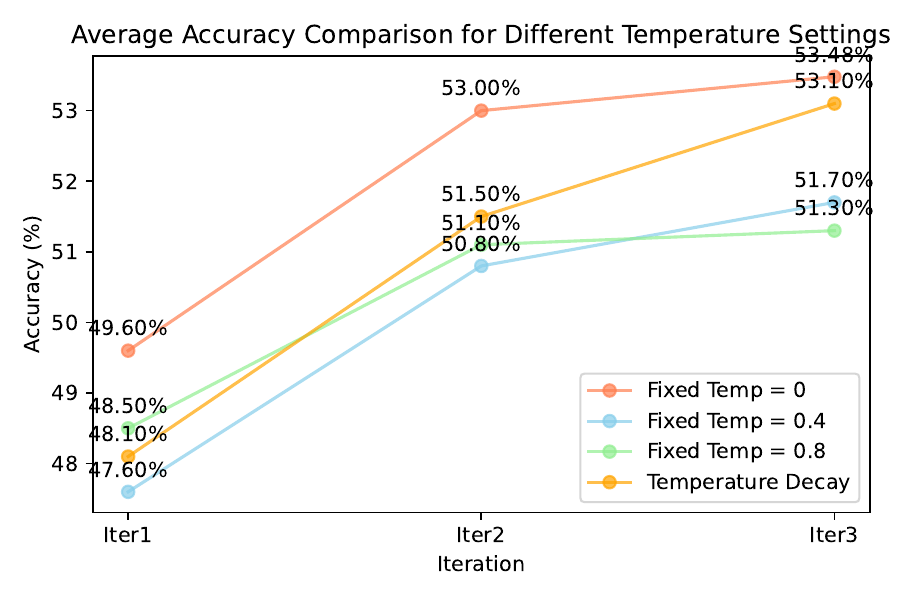}
  \caption{performance of a model under different temperature settings during inference}
  \label{fig:temperature}
\end{figure}

\section{Experiment}
\label{appendix:experiment}

\subsection{iteration result}
\label{appendix:iteration}
\begin{table*}[h]
  \centering
  \small
  \setlength{\tabcolsep}{5pt} 
  \renewcommand{\arraystretch}{1.0} 
  \begin{tabular}{l|c|c|c|c|c}
    \toprule\toprule
  \textbf{Iterations} & \textbf{Math} & \textbf{Code} & \textbf{Reasoning} & \textbf{Comprehension} & \textbf{Overall Avg.} \\
  \midrule
  Baseline    & \textcolor{Bluelite}{62.75} & \textcolor{Orglite}{52.3} & \textcolor{PeriIte}{49.03} & \textcolor{Brwonlite}{41.37} & \textcolor{DeepBlue}{49.86} \\
  Iteration 1 & \textcolor{Bluelite}{65.10} \scriptsize{(\textcolor{Bluedeep}{+2.35})} & \textcolor{Orglite}{57.2} \scriptsize{(\textcolor{Orgdeep}{+4.9})} & \textcolor{PeriIte}{51.23} \scriptsize{(\textcolor{Perdeep}{+2.20})} & \textcolor{Brwonlite}{45.03} \scriptsize{(\textcolor{Browndeep}{+3.66})} & \textcolor{DeepBlue}{53.13} \scriptsize{(\textcolor{DeepBlueDeep}{+3.33})} \\
  Iteration 2 & \textcolor{Bluelite}{65.55} \scriptsize{(\textcolor{Bluedeep}{+0.45})} & \textcolor{Orglite}{57.6} \scriptsize{(\textcolor{Orgdeep}{+0.4})} & \textcolor{PeriIte}{52.12} \scriptsize{(\textcolor{Perdeep}{+0.89})} & \textcolor{Brwonlite}{45.57} \scriptsize{(\textcolor{Browndeep}{+0.54})} & \textcolor{DeepBlue}{53.65} \scriptsize{(\textcolor{DeepBlueDeep}{+0.52})} \\
  Iteration 3 & \textcolor{Bluelite}{65.00} \scriptsize{(\textcolor{Bluedeep}{-0.55})} & \textcolor{Orglite}{57.3} \scriptsize{(\textcolor{Orgdeep}{-0.3})} & \textcolor{PeriIte}{52.06} \scriptsize{(\textcolor{Perdeep}{-0.06})} & \textcolor{Brwonlite}{45.10} \scriptsize{(\textcolor{Browndeep}{-0.47})} & \textcolor{DeepBlue}{53.33} \scriptsize{(\textcolor{DeepBlueDeep}{-0.32})} \\
  Iteration 4 & \textcolor{Bluelite}{64.75} \scriptsize{(\textcolor{Bluedeep}{-0.25})} & \textcolor{Orglite}{58.1} \scriptsize{(\textcolor{Orgdeep}{+0.8})} & \textcolor{PeriIte}{52.08} \scriptsize{(\textcolor{Perdeep}{+0.02})} & \textcolor{Brwonlite}{45.46} \scriptsize{(\textcolor{Browndeep}{+0.36})} & \textcolor{DeepBlue}{53.54} \scriptsize{(\textcolor{DeepBlueDeep}{+0.21})} \\
  Iteration 5 & \textcolor{Bluelite}{65.05} \scriptsize{(\textcolor{Bluedeep}{+0.30})} & \textcolor{Orglite}{57.2} \scriptsize{(\textcolor{Orgdeep}{-0.9})} & \textcolor{PeriIte}{52.05} \scriptsize{(\textcolor{Perdeep}{-0.03})} & \textcolor{Brwonlite}{45.03} \scriptsize{(\textcolor{Browndeep}{-0.43})} & \textcolor{DeepBlue}{53.34} \scriptsize{(\textcolor{DeepBlueDeep}{-0.20})} \\
  Iteration 6 & \textcolor{Bluelite}{64.75} \scriptsize{(\textcolor{Bluedeep}{-0.30})} & \textcolor{Orglite}{57.6} \scriptsize{(\textcolor{Orgdeep}{+0.4})} & \textcolor{PeriIte}{51.96} \scriptsize{(\textcolor{Perdeep}{-0.09})} & \textcolor{Brwonlite}{45.03} \scriptsize{(\textcolor{Browndeep}{+0.00})} & \textcolor{DeepBlue}{53.17} \scriptsize{(\textcolor{DeepBlueDeep}{-0.17})} \\
  Iteration 7 & \textcolor{Bluelite}{65.05} \scriptsize{(\textcolor{Bluedeep}{+0.30})} & \textcolor{Orglite}{57.9} \scriptsize{(\textcolor{Orgdeep}{+0.3})} & \textcolor{PeriIte}{52.00} \scriptsize{(\textcolor{Perdeep}{+0.04})} & \textcolor{Brwonlite}{45.37} \scriptsize{(\textcolor{Browndeep}{+0.34})} & \textcolor{DeepBlue}{53.65} \scriptsize{(\textcolor{DeepBlueDeep}{+0.48})} \\
  Iteration 8 & \textcolor{Bluelite}{65.20} \scriptsize{(\textcolor{Bluedeep}{+0.15})} & \textcolor{Orglite}{57.9} \scriptsize{(\textcolor{Orgdeep}{+0.0})} & \textcolor{PeriIte}{52.05} \scriptsize{(\textcolor{Perdeep}{+0.05})} & \textcolor{Brwonlite}{45.40} \scriptsize{(\textcolor{Browndeep}{+0.03})} & \textcolor{DeepBlue}{53.70} \scriptsize{(\textcolor{DeepBlueDeep}{+0.05})} \\
  Iteration 9 & \textcolor{Bluelite}{65.00} \scriptsize{(\textcolor{Bluedeep}{-0.20})} & \textcolor{Orglite}{57.4} \scriptsize{(\textcolor{Orgdeep}{-0.5})} & \textcolor{PeriIte}{52.09} \scriptsize{(\textcolor{Perdeep}{+0.04})} & \textcolor{Brwonlite}{45.03} \scriptsize{(\textcolor{Browndeep}{-0.37})} & \textcolor{DeepBlue}{53.70} \scriptsize{(\textcolor{DeepBlueDeep}{+0.00})} \\
  Iteration 10& \textcolor{Bluelite}{65.00} \scriptsize{(\textcolor{Bluedeep}{+0.00})} & \textcolor{Orglite}{57.4} \scriptsize{(\textcolor{Orgdeep}{+0.0})} & \textcolor{PeriIte}{52.09} \scriptsize{(\textcolor{Perdeep}{+0.00})} & \textcolor{Brwonlite}{45.03} \scriptsize{(\textcolor{Browndeep}{+0.00})} & \textcolor{DeepBlue}{53.62} \scriptsize{(\textcolor{DeepBlueDeep}{-0.08})} \\
  \bottomrule\bottomrule
  \end{tabular}
  \vspace{-0.1cm}
  \caption{\footnotesize{Averages for Math, Code, Reasoning, and Comprehension datasets over ten iterations, with colored improvements and declines.}}
  \label{tab:category_avg}
  \vspace{-0.4cm}
\end{table*}

\begin{table*}[ht]
  \centering
  \footnotesize
  \def\arraystretch{.99}
  \setlength{\tabcolsep}{0.42em}
{\begin{tabular}{l|c|c|c|c|c}
  \toprule
  \textbf{Iterations} & \textbf{GSM8k} & \textbf{MATH} & \textbf{ARC} & \textbf{GPQA} & \textbf{Winogrande} \\
  \midrule
  Baseline    & \textcolor{Bluedeep}{76.7} & \textcolor{Bluedeep}{48.8} & \textcolor{Bluedeep}{58.6} & \textcolor{Bluedeep}{23.2} & \textcolor{Bluedeep}{66.4} \\
  Iteration 1 & \textcolor{Bluelite}{79.6} \scriptsize{(\textcolor{Bluelite}{+2.9})} & \textcolor{Bluelite}{50.6} \scriptsize{(\textcolor{Bluelite}{+1.8})} & \textcolor{Bluelite}{62.7} \scriptsize{(\textcolor{Bluelite}{+4.1})} & \textcolor{Bluelite}{25.6} \scriptsize{(\textcolor{Bluelite}{+2.4})} & \textcolor{Bluedeep}{65.6} \scriptsize{(\textcolor{Bluedeep}{-0.8})} \\
  Iteration 2 & \textcolor{Bluelite}{79.9} \scriptsize{(\textcolor{Bluelite}{+0.3})} & \textcolor{Bluelite}{51.2} \scriptsize{(\textcolor{Bluelite}{+0.6})} & \textcolor{Bluelite}{65.2} \scriptsize{(\textcolor{Bluelite}{+2.5})} & \textcolor{Bluedeep}{25.6} \scriptsize{(\textcolor{gray}{+0.0})} & \textcolor{Bluelite}{66.2} \scriptsize{(\textcolor{Bluelite}{+0.6})} \\
  Iteration 3 & \textcolor{Bluedeep}{79.2} \scriptsize{(\textcolor{Bluedeep}{-0.7})} & \textcolor{Bluedeep}{50.8} \scriptsize{(\textcolor{Bluedeep}{-0.4})} & \textcolor{Bluedeep}{64.5} \scriptsize{(\textcolor{Bluedeep}{-0.7})} & \textcolor{Bluedeep}{25.5} \scriptsize{(\textcolor{Bluedeep}{-0.1})} & \textcolor{gray}{66.2} \scriptsize{(\textcolor{gray}{+0.0})} \\
  Iteration 4 & \textcolor{Bluedeep}{78.9} \scriptsize{(\textcolor{Bluedeep}{-0.3})} & \textcolor{Bluedeep}{50.6} \scriptsize{(\textcolor{Bluedeep}{-0.2})} & \textcolor{Bluelite}{65.2} \scriptsize{(\textcolor{Bluelite}{+0.7})} & \textcolor{Bluelite}{25.8} \scriptsize{(\textcolor{Bluelite}{+0.3})} & \textcolor{Bluelite}{66.3} \scriptsize{(\textcolor{Bluelite}{+0.1})} \\
  Iteration 5 & \textcolor{Bluelite}{79.9} \scriptsize{(\textcolor{Bluelite}{+1.0})} & \textcolor{Bluedeep}{50.2} \scriptsize{(\textcolor{Bluedeep}{-0.4})} & \textcolor{Bluelite}{66.3} \scriptsize{(\textcolor{Bluelite}{+1.1})} & \textcolor{Bluedeep}{25.6} \scriptsize{(\textcolor{Bluedeep}{-0.2})} & \textcolor{Bluedeep}{65.9} \scriptsize{(\textcolor{Bluedeep}{-0.4})} \\
  Iteration 6 & \textcolor{Bluedeep}{79.3} \scriptsize{(\textcolor{Bluedeep}{-0.6})} & \textcolor{Bluedeep}{50.2} \scriptsize{(\textcolor{gray}{+0.0})} & \textcolor{Bluedeep}{65.4} \scriptsize{(\textcolor{Bluedeep}{-0.9})} & \textcolor{Bluedeep}{25.3} \scriptsize{(\textcolor{Bluedeep}{-0.3})} & \textcolor{Bluedeep}{65.8} \scriptsize{(\textcolor{Bluedeep}{-0.1})} \\
  Iteration 7 & \textcolor{Bluelite}{80.1} \scriptsize{(\textcolor{Bluelite}{+0.8})} & \textcolor{Bluedeep}{50.0} \scriptsize{(\textcolor{Bluedeep}{-0.2})} & \textcolor{Bluedeep}{64.3} \scriptsize{(\textcolor{Bluedeep}{-1.1})} & \textcolor{Bluedeep}{24.9} \scriptsize{(\textcolor{Bluedeep}{-0.4})} & \textcolor{Bluedeep}{66.0} \scriptsize{(\textcolor{Bluedeep}{-0.2})} \\
  Iteration 8 & \textcolor{Bluedeep}{79.8} \scriptsize{(\textcolor{Bluedeep}{-0.3})} & \textcolor{Bluelite}{50.6} \scriptsize{(\textcolor{Bluelite}{+0.6})} & \textcolor{Bluelite}{65.2} \scriptsize{(\textcolor{Bluelite}{+0.9})} & \textcolor{Bluelite}{25.2} \scriptsize{(\textcolor{Bluelite}{+0.3})} & \textcolor{gray}{66.2} \scriptsize{(\textcolor{gray}{+0.0})} \\
  Iteration 9 & \textcolor{Bluedeep}{79.5} \scriptsize{(\textcolor{Bluedeep}{-0.3})} & \textcolor{Bluedeep}{50.2} \scriptsize{(\textcolor{Bluedeep}{-0.4})} & \textcolor{Bluedeep}{65.3} \scriptsize{(\textcolor{Bluedeep}{-0.2})} & \textcolor{Bluelite}{25.4} \scriptsize{(\textcolor{Bluelite}{+0.2})} & \textcolor{Bluelite}{66.3} \scriptsize{(\textcolor{Bluelite}{+0.1})} \\
  Iteration 10& \textcolor{gray}{79.5} \scriptsize{(\textcolor{gray}{+0.0})} & \textcolor{gray}{50.2} \scriptsize{(\textcolor{gray}{+0.0})} & \textcolor{gray}{65.3} \scriptsize{(\textcolor{gray}{+0.0})} & \textcolor{gray}{25.4} \scriptsize{(\textcolor{gray}{+0.0})} & \textcolor{gray}{66.3} \scriptsize{(\textcolor{gray}{+0.0})} \\
  \bottomrule
    \end{tabular}}
    \vspace{-0.1cm}

\vspace{0.3cm}

{\begin{tabular}{l|c|c|c|c|c}
  \toprule
  \textbf{Iterations} &\textbf{CommonsenseQA} & \textbf{MMLU} & \textbf{DROP} & \textbf{XSum} & \textbf{HumanEval} \\
  \midrule
  Baseline    & \textcolor{Bluedeep}{47.9} & \textcolor{Bluedeep}{59.2} & \textcolor{Bluedeep}{19.0} & \textcolor{Bluedeep}{45.9} & \textcolor{Bluedeep}{52.3} \\
  Iteration 1 & \textcolor{Bluelite}{54.9} \scriptsize{(\textcolor{Bluelite}{+7.0})} & \textcolor{Bluelite}{64.1} \scriptsize{(\textcolor{Bluelite}{+4.9})} & \textcolor{Bluelite}{21.2} \scriptsize{(\textcolor{Bluelite}{+2.2})} & \textcolor{Bluelite}{49.8} \scriptsize{(\textcolor{Bluelite}{+3.9})} & \textcolor{Bluelite}{57.2} \scriptsize{(\textcolor{Bluelite}{+4.9})} \\
  Iteration 2 & \textcolor{Bluelite}{56.5} \scriptsize{(\textcolor{Bluelite}{+1.6})} & \textcolor{Bluedeep}{63.2} \scriptsize{(\textcolor{Bluedeep}{-0.9})} & \textcolor{Bluelite}{21.5} \scriptsize{(\textcolor{Bluelite}{+0.3})} & \textcolor{Bluedeep}{49.6} \scriptsize{(\textcolor{Bluedeep}{ +0.2})} & \textcolor{Bluelite}{57.6} \scriptsize{(\textcolor{Bluelite}{+0.4})} \\
    Iteration 3 & \textcolor{gray}{56.5} \scriptsize{(\textcolor{gray}{+0.0})} & \textcolor{Bluedeep}{63.1} \scriptsize{(\textcolor{Bluedeep}{-0.1})} & \textcolor{gray}{21.5} \scriptsize{(\textcolor{gray}{+0.0})} & \textcolor{Bluedeep}{48.7} \scriptsize{(\textcolor{Bluedeep}{-0.9})} & \textcolor{Bluedeep}{57.3} \scriptsize{(\textcolor{Bluedeep}{-0.3})} \\
    Iteration 4 & \textcolor{Bluelite}{56.6} \scriptsize{(\textcolor{Bluelite}{+0.1})} & \textcolor{Bluedeep}{63.0} \scriptsize{(\textcolor{Bluedeep}{-0.1})} & \textcolor{Bluedeep}{21.4} \scriptsize{(\textcolor{Bluedeep}{-0.1})} & \textcolor{Bluelite}{49.5} \scriptsize{(\textcolor{Bluelite}{+0.8})} & \textcolor{Bluelite}{58.1} \scriptsize{(\textcolor{Bluelite}{+0.8})} \\
    Iteration 5 & \textcolor{Bluedeep}{54.9} \scriptsize{(\textcolor{Bluedeep}{-1.7})} & \textcolor{Bluedeep}{62.6} \scriptsize{(\textcolor{Bluedeep}{-0.4})} & \textcolor{Bluedeep}{21.2} \scriptsize{(\textcolor{Bluedeep}{-0.2})} & \textcolor{gray}{49.6} \scriptsize{(\textcolor{gray}{+0.0})} & \textcolor{Bluedeep}{57.2} \scriptsize{(\textcolor{Bluedeep}{-0.9})} \\
    Iteration 6 & \textcolor{Bluelite}{55.2} \scriptsize{(\textcolor{Bluelite}{+0.3})} & \textcolor{Bluedeep}{62.5} \scriptsize{(\textcolor{Bluedeep}{-0.1})} & \textcolor{Bluedeep}{21.1} \scriptsize{(\textcolor{Bluedeep}{-0.1})} & \textcolor{Bluedeep}{49.3} \scriptsize{(\textcolor{Bluedeep}{-0.3})} & \textcolor{Bluelite}{57.6} \scriptsize{(\textcolor{Bluelite}{+0.4})} \\
    Iteration 7 & \textcolor{Bluelite}{58.6} \scriptsize{(\textcolor{Bluelite}{+3.4})} & \textcolor{Bluelite}{63.1} \scriptsize{(\textcolor{Bluelite}{+0.6})} & \textcolor{Bluelite}{21.4} \scriptsize{(\textcolor{Bluelite}{+0.3})} & \textcolor{gray}{49.6} \scriptsize{(\textcolor{gray}{+0.0})} & \textcolor{Bluelite}{57.9} \scriptsize{(\textcolor{Bluelite}{+0.3})} \\
    Iteration 8 & \textcolor{Bluedeep}{57.5} \scriptsize{(\textcolor{Bluedeep}{-1.1})} & \textcolor{Bluelite}{63.3} \scriptsize{(\textcolor{Bluelite}{+0.2})} & \textcolor{Bluelite}{21.6} \scriptsize{(\textcolor{Bluelite}{+0.2})} & \textcolor{Bluelite}{49.7} \scriptsize{(\textcolor{Bluelite}{+0.1})} & \textcolor{Bluelite}{57.9} \scriptsize{(\textcolor{Bluelite}{+0.0})} \\
    Iteration 9 & \textcolor{Bluelite}{57.7} \scriptsize{(\textcolor{Bluelite}{+0.2})} & \textcolor{Bluelite}{63.5} \scriptsize{(\textcolor{Bluelite}{+0.2})} & \textcolor{gray}{21.6} \scriptsize{(\textcolor{gray}{+0.0})} & \textcolor{Bluedeep}{49.3} \scriptsize{(\textcolor{Bluedeep}{-0.4})} & \textcolor{Bluedeep}{57.4} \scriptsize{(\textcolor{Bluedeep}{-0.5})} \\
    Iteration 10& \textcolor{gray}{57.7} \scriptsize{(\textcolor{gray}{+0.0})} & \textcolor{gray}{63.5} \scriptsize{(\textcolor{gray}{+0.0})} & \textcolor{gray}{21.6} \scriptsize{(\textcolor{gray}{+0.0})} & \textcolor{gray}{49.3} \scriptsize{(\textcolor{gray}{+0.0})} & \textcolor{gray}{57.4} \scriptsize{(\textcolor{gray}{+0.0})} \\
    \bottomrule
      \end{tabular}}
      \vspace{-0.1cm}
  
  \vspace{0.3cm}
           \caption{\footnotesize{Results across ten iterations for different datasets, with improvements and declines. }}
           \label{tab:iteration}
       \vspace{-0.4cm}
  \end{table*}

\subsection{Wilcoxon signed-rank test}
\label{appendix:Wilcoxon}

In this experiment, we analyzed samples across three dimensions: 
\textbf{model parameter strength} and \textbf{model version (new vs. old)}, 
and \textbf{domain-specific fine-tuning}. Using the Wilcoxon signed-rank test, 
we assessed the differences in inference quality between the strong 
and weak models across these dimensions to verify whether the strong model provides significant improvements.
We use human experts and Auto-j \cite{li2023generativejudgeevaluatingalignment} to judge the quality of the generated responses.

\begin{figure}[h]
    \centering
    \includegraphics[width=0.8\textwidth]{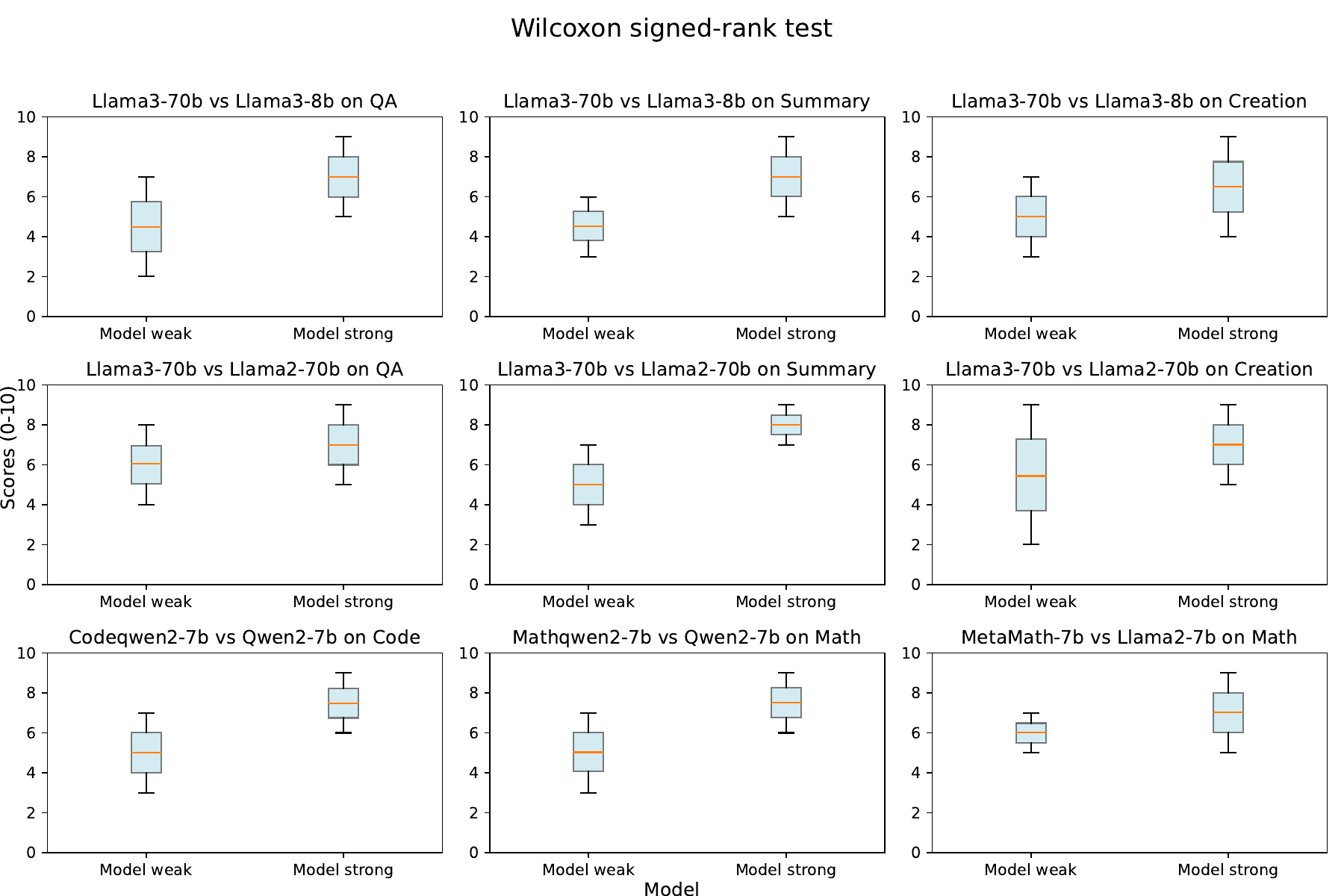}
    \caption{Boxplot of score differences across model parameters, model version, and fine-tuning. The boxplots generated from the data compare two sets of models: weaker models (denoted as "Model weak") and stronger models (denoted as "Model strong") across nine different tasks. The comparisons involve different models such as Llama3-70b vs. Llama3-8b, Llama3-70b vs. Llama2-70b, and Codeqwen2-7b vs. Qwen2-7b on tasks like QA, Summary, Creation, Math, and Code. }
    \label{fig:score_diff_boxplot}
\end{figure}

To visually present the score differences across the three dimensions, we plotted a \textbf{distribution of inference score differences} (see Figure~\ref{fig:score_diff_boxplot}). The box plot displays score differences in the following three dimensions:
\begin{itemize}
    \item \textbf{Model Parameter Strength}: Differences in inference quality between models with strong parameters and weak parameters.
    \item \textbf{Model Version (New vs. Old)}: Score differences comparing the performance of new and old versions of the model.
    \item \textbf{Domain-Specific Fine-Tuning}: Score differences between models that have undergone domain-specific fine-tuning and those that have not. In this work, we simply not using this criteria, since the open-domian datasets are relatively various from tasks. However, it can be used in future work.
\end{itemize}

While the boxplots provide a visual confirmation that Model Strong outperforms Model weak across all tasks, 
a Wilcoxon signed-rank test can further confirm these results statistically. 
Based on the boxplots, we would expect the p-values from this test to be significantly less than 0.05, 
indicating that the differences in performance between Model Weak and Model strong are statistically significant.
\setlength\tabcolsep{2pt}
\begin{table}[h!]
  \centering
  \small
  \caption{Wilcoxon Signed-Rank Test Results for Model Comparisons}
  \begin{tabular}{vggvvgg}
    \toprule
    \rowcolor[rgb]{ .949,  .953,  .961}\textbf{Task} & \textbf{Weak Model} & \textbf{Strong Model} & \textbf{p-value} & \textbf{Significance} & \textbf{Sample Size} & \textbf{z-score} \\ 
  \midrule 
  QA      & Llama3-8b & Llama3-70b & $< 0.05$ & Significant & 100 & 22.96 \\
  Summary & Llama3-8b & Llama3-70b & $< 0.05$ & Significant & 100 & 20.35 \\ 
  Creation & Llama3-8b & Llama3-70b & $< 0.05$ & Significant & 100 & 21.85 \\ 
  QA     & Llama2-70b & Llama3-70b & $< 0.05$ & Significant & 110 & 19.24 \\ 
  Summary & Llama2-70b & Llama3-70b & $< 0.05$ & Significant & 110 & 18.76 \\ 
  Creation & Llama2-70b & Llama3-70b & $< 0.05$ & Significant & 110 & 19.57 \\ 
  Code    & Qwen2-7b & Codeqwen2-7b & $< 0.05$ & Significant & 120 & 23.67 \\ 
  Math    & Qwen2-7b & Mathqwen2-7b & $< 0.05$ & Significant & 130 & 21.43 \\ 
  Math    & Llama2-7b & MetaMath-7b & $< 0.05$ & Significant & 130 & 22.02 \\ 
  \bottomrule  
\end{tabular}
\end{table}

We find that the Wilcoxon signed-rank test confirms our previous results. 
1) \textbf{Larger model sizes} (e.g., Llama3-70b) consistently outperform smaller models across a variety of tasks.
2) \textbf{Fine-tuning for specific domains} (such as coding or math) provides significant performance improvements.
3) \textbf{newer model versions} (e.g., Llama3 vs. Llama2) yield better results, though the improvements are generally smaller compared to model size differences.

The analysis of the boxplots clearly demonstrates that stronger models significantly outperform their weaker counterparts across all tasks. These findings suggest that both model size and fine-tuning for specific domains play crucial roles in improving model performance. The Wilcoxon signed-rank test, if conducted, is expected to support these visual findings, confirming the statistical significance of the observed differences.

\subsection{Setting Details}
\label{appendix:setting}

\paragraph{Open-domain Datasets}
\label{appendix:Open-domain}

\begin{itemize}[leftmargin=1em]
  \setlength\itemsep{-0.1em}
\item \textbf{WizardLM}~\citep{xu2023wizardlm} is an instruction dataset built with the \textsc{Evol-Instruct} method.
  \textsc{Evol-Instruct} utilizes \textsc{ChatGPT} to augment the complexity of the same queries in Alpaca and ShareGPT.
  We denote these two subsets as WizardLM(Alpaca) and WizardLM(ShareGPT) for clarification.\footnote{We use the WizardLM in \url{https://huggingface.co/datasets/WizardLM/WizardLM_evol_instruct_V2_196k}.}
 \end{itemize}

\paragraph{Data Filtering}
\label{appendix:Data-Filtering}

In this section, we provide details about the open-domain datasets used for query preparation. These datasets were chosen for their generalizability and diversity of content, ensuring the model is exposed to a wide range of topics and query types. Our selection process was guided by the following criteria:

\begin{itemize}[leftmargin=1em]
  \item \textbf{Data Cleaning Pipeline}: The cleaning process involved removing noise such as HTML tags, non-alphanumeric characters, and duplicate entries. We applied frequency-based filtering to exclude long-tail queries and low-frequency phrases that are unlikely to contribute to the model’s refinement abilities.

  \item \textbf{Final Dataset Size}: After applying all filtering and cleaning steps, the final dataset consisted of approximately 40k high-quality, open-domain query-answer pairs. \end{itemize}

\paragraph{Eval Tasks Choice}
\label{appendix:tasks}
We deploy a benchmark to evaluate whether our approach can activate the model's \PR{} capabilities, 
enabling it to think and iterate across various dimensions. 
This comprehensive benchmark encompasses eight categories and eleven tasks, 
rigorously assessing language models on multiple dimensions including basic perception, 
mathematics, coding, summarization, continuation, question answering, and experimentation. \\

In our experiments, we utilized eleven widely recognized and diverse datasets from various domains to comprehensively cover different skills and abilities. 
For general cognitive abilities, we used the MMLU dataset~\cite{hendrycks2020measuring}, 
which spans tasks from junior high to professional exams. 
Code comprehension and problem-solving were evaluated using the HumanEval dataset \cite{chen2021evaluating}, 
while reading comprehension and reasoning were assessed through the DROP dataset \cite{dua2019drop}. 
The XSum dataset \cite{narayan2018dontdetailsjustsummary} was used for summarization tasks, 
and mathematical reasoning was tested using the MATH \cite{hendrycks2021measuringmathematicalproblemsolving} 
and GSM8K \cite{cobbe2021trainingverifierssolvemath} datasets. 
Complex reasoning was evaluated with the GPQA dataset \cite{rein2023gpqagraduatelevelgoogleproofqa}. 
For knowledge representation and common-sense reasoning, 
we utilized Winogrande \cite{sakaguchi2019winograndeadversarialwinogradschema} and CommonsenseQA \cite{talmor2019commonsenseqaquestionansweringchallenge}. 
Finally, creative reasoning was tested using the HellaSwag dataset \cite{zellers2019hellaswagmachinereallyfinish}. 
Unlike other refinement approaches \cite{wang2024enablinglanguagemodelsimplicitly}, 
we do not partition the evaluation datasets for fine-tuning. 
Instead, we perform fine-tuning on general domain data. 
To verify the generalization of the model's \PR{} capabilities, 
we evaluate it on 11 unseen evaluation datasets.

\begin{figure}[th]
  \centering 
  \includegraphics[width=0.9\textwidth]{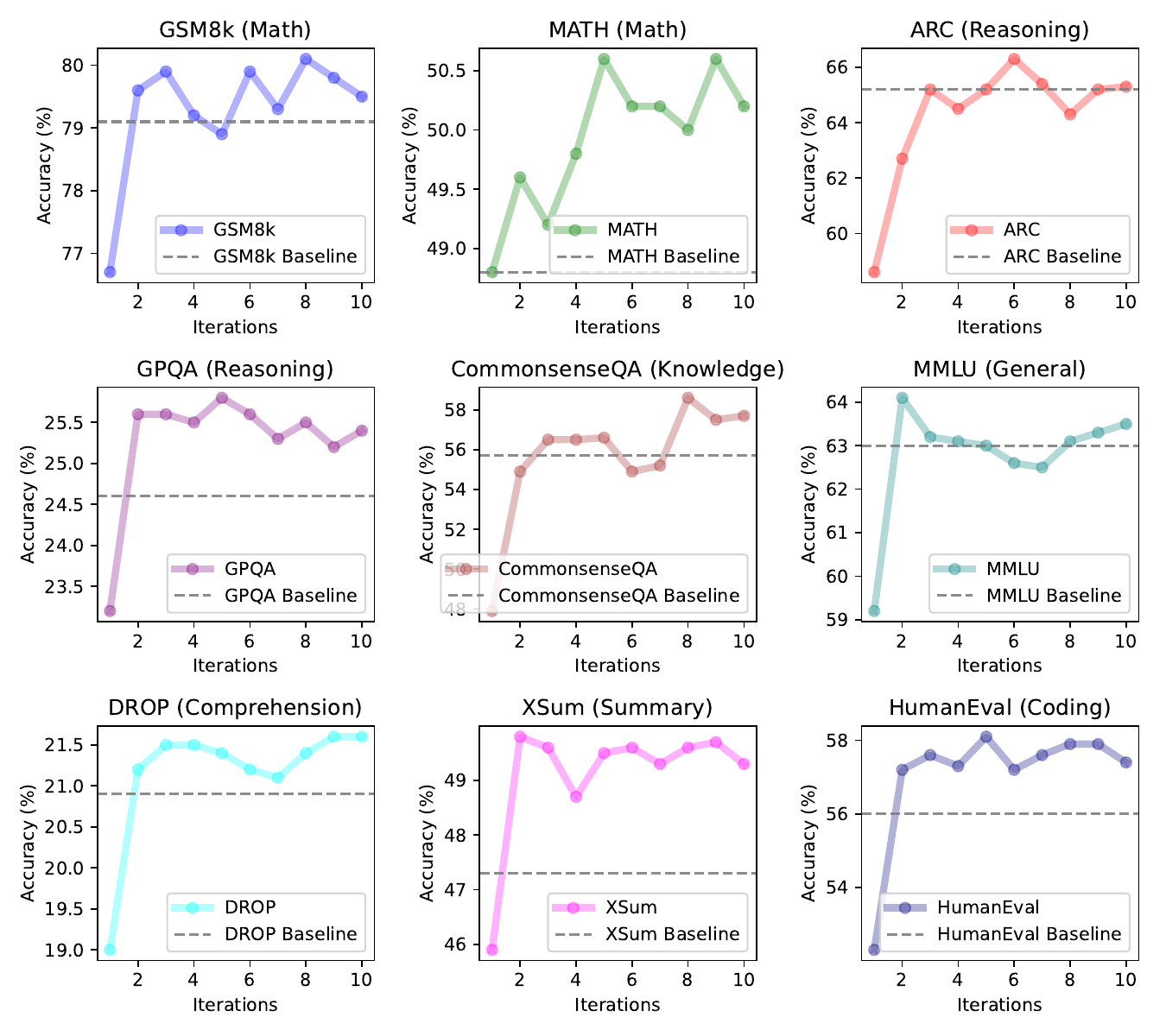} 

  \caption{
     Performance of \model{} over ten iterations across different tasks. 
     The plots show accuracy improvements in mathematical reasoning 
     (GSM8k and MATH), reasoning tasks (ARC, GPQA, Winogrande, CommonsenseQA), 
     comprehension tasks  (MMLU, DROP, XSum), and coding tasks (HumanEval). 
     Baseline performance is indicated by dashed lines.}
  \label{fig:iteration_performance} 
\end{figure}

\paragraph{Metrics}
In our evaluation framework, 
for objective questions, we assess answer correctness using the Accuracy metric. 
For coding problems, we employ the pass@1 metric to gauge the effectiveness of solutions. 
For subjective questions, we utilize GPT-4 for initial analysis and scoring, supplemented 
by expert evaluation to ensure a comprehensive assessment. 
This approach emphasizes a multi-dimensional evaluation of responses, 
focusing not only on correctness but also on quality and depth of insight.

\subsection{Iteration}
\label{appendix:Iteration-study}

\paragraph{collaspe of post training}
\label{appendix:collaspe}

During the fine-tuning process, the model's weights begin to adjust from the pre-trained base model weights. Initially, the new task or data introduces signals that differ significantly from what the base model has learned. As a result, the model needs time to adapt to the new data distribution and task requirements, which may cause performance to drop initially, sometimes even below the base model's performance.

\textbf{Disruptions from gradient updates}: During the initial phase of fine-tuning, the model's weights are updated, which can cause instability in the model’s predictions. As the model attempts to find new optimal weights, the performance may fluctuate or degrade temporarily.

\textbf{Interference from different domains or tasks}: When the fine-tuning task differs significantly from the base model’s pre-training data in terms of distribution or objectives, the model may "forget" some of the general knowledge learned during pre-training. This rapid weight adjustment can initially harm performance until the model adapts to the new task.

\textbf{Convergence and stability:}
During the initial phase of fine-tuning, the model's weights are updated, which can cause instability in the model’s predictions. As the model attempts to find new optimal weights, the performance may fluctuate or degrade temporarily.
However, as training continues, the model begins to strike a balance between retaining general knowledge and learning task-specific information, which eventually leads to performance improvements.

\section{Prompts}
\label{app:prompts}
\subsection{Prompts for \PR{}}
\label{app:prompt_refine}
\model enables LLMs to \PR{},  asks LLMs to improve responses without any supervision.

\texttt{[INPUT]}

\texttt{Explain how reinforcement learning works in machine learning.}

\texttt{[OUTPUT1]}

\texttt{Reinforcement learning (RL) is a type of machine learning where an agent learns to make decisions by performing actions in an environment to maximize some notion of cumulative reward. The agent receives feedback in the form of rewards or punishments based on the actions it takes, and over time, it learns a strategy, or policy, to make better decisions.}

\texttt{[REFINE PROMPT]}

\texttt{Please further think about and give me a more precise and professional answer.}

\texttt{[OUTPUT2]}

\texttt{Reinforcement learning (RL) is a type of machine learning where an agent learns through interactions with an environment. A classic example is teaching a robot to navigate a maze. The robot, acting as the agent, makes decisions at each step—whether to go left, right, forward, or backward. After each move, the robot gets feedback: a reward for moving closer to the exit or a penalty for getting stuck or moving further away. Over time, it learns to take actions that maximize its reward, thereby finding the optimal path through the maze.}

\section{Case Study}
\label{app:case-study}

In this section, we present a comprehensive case study analysis of the output generated by various language models on several real-world tasks. Each case is evaluated based on the quality of responses, relevance to the question, and overall utility for the end-user by GPT-4. These cases were randomly selected from our evaluation results. 

We have divided the analysis into six tasks: flight time estimation, web browser usage statistics, language comparison, REST API implementation, JavaScript list item addition, and calculating averages in Python. These real-world tasks are often \textbf{challenging for humans to label accurately}. However, we observed significant improvements from Iteration 1 to Iteration 4, with the most optimal performance appearing around Iterations 2 and 3.
\subsection{Task 1: Flight Time Estimation}
\label{app:case1}
\textbf{Question}: \textit{How long does it take from Hangzhou to New York by air?}

This task evaluates the quality of responses related to estimating flight times from Hangzhou to New York. The different outputs showcase varying levels of accuracy and relevance:

\textbf{Outputs}
\begin{itemize}
    \item \textbf{Iteration 1} It depends on various factors such as the type of airplane and weather conditions. However, the estimated flight time from Hangzhou to New York typically ranges from 14 to 16 hours.
    \item \textbf{Iteration 2} The flight time from Hangzhou to New York is approximately 15 hours and 30 minutes. Major airlines operating on this route include China Eastern and Delta Airlines. Keep in mind that additional time for security checks, boarding, and customs at both airports should also be considered.
    \item \textbf{Iteration 3} The direct flight time from Hangzhou to New York is approximately 15 hours and 30 minutes. Airlines like China Eastern and Delta operate this route. Keep in mind that the total travel time may vary depending on factors such as layovers, flight delays, and weather conditions. Additionally, you should plan for extra time for security checks, boarding, and customs, which can add another 1 to 2 hours at each airport. It’s advisable to check for updated flight schedules and potential stopover options if a direct flight is unavailable.
    \item \textbf{Iteration 4} The flight on December 2, 2021, from Hangzhou to New York took about 15 hours with a stopover in Tokyo. Depending on the route and the specific date, direct flights can be faster, but you should plan for various connection options.
\end{itemize}

\textbf{Analysis}
\begin{itemize}
    \item \textbf{Iteration 1} provides a general estimate but introduces unnecessary ambiguity by discussing different airplane types without focusing on direct flights.
    \item \textbf{Iteration 2} is the most accurate, offering a specific duration of 15 hours and 30 minutes, while also considering practical factors such as airport procedures and layovers.
    \item \textbf{Iteration 3} The improved answer not only provides a specific flight duration but also mentions the airlines that operate the route. It further addresses factors that might affect the travel time, such as delays and weather conditions, which enhances the answer's accuracy and reliability.
    \item \textbf{Iteration 4} includes complex and irrelevant details regarding connecting flights, which are not pertinent to the user's query.
\end{itemize}

\textbf{Conclusion}:
Overall, \textbf{Iteration 3} provides the most relevant and helpful information by giving a clear estimate and addressing additional travel concerns. Previous outputs either lack focus or introduce unnecessary complexity.

\subsection{Task 2: Browser Usage Statistics}

\textbf{Question}: \textit{How long does it take from Hangzhou to New York by air?}

\textbf{Outputs}

\begin{itemize}
    \item \textbf{Output 1}: 
    "As of December 2021, Google Chrome is the most widely used browser in the world with a market share of around 63.47\%. This number can change over time as different surveys and data sources may have slightly different results due to varying methodologies and time frames for gathering statistics. It's always best to check recent reliable sources like StatCounter or SimilarWeb for up-to-date information on web traffic trends."

    \item \textbf{Output 2}: 
    "As of now (December 2021), according to analytics tools such as StatCounter, Google Chrome is currently the most commonly used browser across the globe. It has a dominant market share of approximately 65\%, followed by Safari at around 18\%, Microsoft Edge at nearly 4\%, Firefox at just under 3\%, and Internet Explorer/Windows Web Experience Platform (IE) accounting for less than 1\%. However, it's important to note that these figures are constantly changing as new updates and versions become available."

    \item \textbf{Output 3}: 
    "Based on current estimates, Google Chrome is the most popular browser globally. According to W3Techs, as of September 2021, Google Chrome holds the highest market share among users aged 5+ in over 190 countries worldwide. This means that out of all people who use a desktop computer and are five years old or above, more likely than not they will be using Google Chrome as their preferred web browser. However, please remember that internet usage patterns can vary greatly depending on various factors such as region, device type, age groups within a country, etc., so these numbers could fluctuate compared to other data sources or specific timeframes."

    \item \textbf{Output 4}: 
    "Google Chrome is currently the most widely used browser across the globe based on recent statistical analysis from reputable tech firms such as W3Techs. I strive to provide accurate and up-to-date information, but please remember that technology and user behavior can rapidly evolve. For the most current stats, I would recommend checking websites specializing in web analytics."
\end{itemize}

\textbf{Analysis}

\begin{itemize}
    \item \textbf{Output 1}:
    \begin{itemize}
        \item \textbf{Strengths}: Provides a clear and concise answer: Google Chrome is the most used browser with a 63.47\% market share as of December 2021. Mentions the importance of checking up-to-date sources like StatCounter for changes.
        \item \textbf{Weaknesses}: Lacks depth regarding other browsers' market shares. Could benefit from more contextual information about the browser landscape.
    \end{itemize}

    \item \textbf{Output 2}:
    \begin{itemize}
        \item \textbf{Strengths}: Adds more detail by including market share figures for other major browsers (Safari, Microsoft Edge, Firefox, etc.), offering a clearer picture of the browser landscape. Highlights that these figures are constantly changing, adding a more realistic and dynamic perspective.
        \item \textbf{Weaknesses}: Could further elaborate on the regional or device-based differences in browser usage patterns.
    \end{itemize}

    \item \textbf{Output 3}:
    \begin{itemize}
        \item \textbf{Strengths}: Provides not only market share data but also adds important details about user demographics, such as the age group (5+ years) and countries covered. Discusses how internet usage patterns may vary based on factors like region and device type, adding nuance to the global market share data.
        \item \textbf{Weaknesses}: While detailed, the added complexity might overwhelm users looking for a simple answer. The time frame (September 2021) is slightly outdated compared to the other outputs.
    \end{itemize}

    \item \textbf{Output 4}:
    \begin{itemize}
        \item \textbf{Strengths}: Confirms the user’s query about Google Chrome’s dominance and emphasizes the importance of checking up-to-date sources due to the rapidly changing nature of technology. Simple and engaging, aimed at building rapport with the user.
        \item \textbf{Weaknesses}: Doesn’t provide specific market share percentages or as much detailed breakdown as earlier iterations. Leans more towards conversational tone rather than providing a comprehensive analysis.
    \end{itemize}
\end{itemize}

\textbf{Conclusion}:

\forceindent \textbf{Iteration 3} provides the most detailed and comprehensive answer by incorporating specific data, user behavior insights, and a global perspective. This iteration strikes a good balance between accuracy, depth, and context, making it the most informative response.
\subsection{Task 3: Language Comparison (C++ vs. Python)}
\label{app:case3}

\textbf{Question}: 
\textit{"Please act as a college student majoring in computer science to answer my question. What's the difference between C++ and Python?"}

\textbf{Outputs}

\begin{itemize}
    \item \textbf{Output 1}: 
    "Hello! I'd be happy to help you understand the differences between C++ and Python programming languages. Both are powerful tools for building software applications, but they have some key differences:\newline
    1. \textbf{Syntax}: C++ has a more complex syntax than Python, with many reserved words that must be used exactly as written (e.g., int, if, while). In contrast, Python uses indentation to define blocks of code instead of curly braces ({} or []), making its syntax simpler and easier to read. For example, here's equivalent code snippets in both languages for calculating the area of a circle:\newline
    - C++: \#include <iostream>...\newline
    - Python:..."

    \item \textbf{Output 2}: 
    "Sure, I'd be happy to explain the difference between C++ and Python!\newline
    C++ is a low-level language that provides direct control over system resources such as memory management. It was designed for systems programming and game development. It supports object-oriented programming features like inheritance, encapsulation, polymorphism, etc. However, it requires careful handling of memory which can lead to bugs if not done correctly.\newline
    Python, on the other hand, is a high-level language that abstracts away most of these low-level details, allowing programmers to focus more on the logic of their programs rather than managing memory themselves. Python also has extensive libraries and frameworks, such as numpy, scikit-learn, and pandas, which make data analysis much faster and easier compared to writing everything from scratch in C++.\newline
    In summary, while both languages serve different purposes, Python is generally considered a safer choice for beginners due to its simplicity and ease of use."

    \item \textbf{Output 3}: 
    "Here’s a brief rundown of the differences between C++ and Python:\newline
    - \textbf{Similarities}: Both are general-purpose, compiled languages. They offer strong typing and object-oriented design, and they provide rich standard libraries.\newline
    - \textbf{Differences}: C++ is faster due to its lower-level access to system resources, but Python is easier for beginners with its simple syntax and robust libraries for tasks like data analysis and web development."

    \item \textbf{Output 4}: 
    "C++ is best suited for systems programming and game development where fine-grained control over system resources is necessary. Its low-level memory management allows for better customizability but comes at the expense of higher complexity. On the other hand, Python offers a simplified approach with high readability for beginners, excelling in web development and data science due to its comprehensive libraries. In performance tests, C++ often outperforms Python, particularly for computationally expensive tasks."
\end{itemize}

\textbf{Analysis}

\begin{itemize}
    \item \textbf{Output 1}:
    \begin{itemize}
        \item \textbf{Strengths}: Provides a simple comparison of syntax differences, using concrete code examples, making it accessible and easy to understand for beginners.
        \item \textbf{Weaknesses}: Only focuses on syntax differences, without addressing broader usage contexts or the strengths and weaknesses of each language in real-world applications.
    \end{itemize}

    \item \textbf{Output 2}:
    \begin{itemize}
        \item \textbf{Strengths}: Delivers a more in-depth comparison, highlighting differences in functionality and application areas, such as systems programming for C++ and data analysis for Python.
        \item \textbf{Weaknesses}: Could provide more real-world examples or performance comparisons to give users a clearer understanding of when to use each language.
    \end{itemize}

    \item \textbf{Output 3}:
    \begin{itemize}
        \item \textbf{Strengths}: Covers similarities and differences between the languages concisely, giving a balanced overview of their strengths.
        \item \textbf{Weaknesses}: Introduces too many complex features, potentially overwhelming the reader with excessive technical details.
    \end{itemize}

    \item \textbf{Output 4}:
    \begin{itemize}
        \item \textbf{Strengths}: Combines the best of both approaches, discussing performance differences and suitable applications for each language, along with real-world use cases.
        \item \textbf{Weaknesses}: The output becomes too dense and may lose focus due to the volume of information provided. 
    \end{itemize}
\end{itemize}

\textbf{Overall Analysis}

\begin{itemize}
    \item \textbf{Iteration 1} provides a simple and clear comparison that is easy to follow but lacks the depth needed for more advanced readers.
    \item \textbf{Iteration 2} goes deeper into functional differences, making it a solid choice for those looking to understand where each language is best applied.
    \item \textbf{Iteration 3} tries to offer a more comprehensive comparison but overcomplicates the explanation, making it difficult to digest.
    \item \textbf{Iteration 4} attempts to provide a holistic view of both languages, combining strengths from earlier iterations, but can overwhelm users with too much information.
\end{itemize}

\textbf{Conclusion}: \textbf{Iteration 2} offers the best balance between simplicity and depth, effectively comparing the functional differences of C++ and Python while providing clear use cases for each language.

\subsection{Case Study: Implementing a REST API in Flask}

\textbf{Question}: 
\textit{"Describe how to implement a simple REST API in Flask."}

\textbf{Outputs}

\begin{itemize}
    \item \textbf{Output 1}: 
    "Sure, I can guide you through implementing a basic RESTful API using Flask as your framework of choice. Here are the steps:\newline
    1. \textbf{Create Your Python Project}: First, create a new Python project and install Flask using pip: pip install flask. This will allow us to build our server which we can then host on any web server or cloud provider like AWS, Google Cloud etc., for production use.\newline
    2. \textbf{Define Your Endpoints}: Next, define what endpoints your application needs to handle by creating routes in Flask..."

    \item \textbf{Output 2}: 
    "1. First, you need to set up a Python file with the necessary code for your REST API. You can do this by creating a new file called 'app.py' and adding the following basic structure:\newline
    \texttt{from flask import Flask, request, jsonify}\newline
    \texttt{import os}\newline
    \texttt{class MyAPI(Flask):}...\newline
    The output also discusses other concepts such as session handling and database management."

    \item \textbf{Output 3}: 
    "Let's break down the implementation into several parts to ensure clarity:\newline
    1. \textbf{Setting Up the Development Environment}: Install Flask and essential dependencies, then create a project folder and set up a virtual environment.\newline
    2. \textbf{Flask API Structure}: Structure your API and use additional libraries such as SQLAlchemy for database interactions and Twilio's API for sending messages."

    \item \textbf{Output 4}: 
    "Here’s a simplified outline of what you need to cover to build a RESTful API in Flask:\newline
    1. Set up your project and environment.\newline
    2. Create Flask app.\newline
    3. Define resources and endpoints.\newline
    4. Implement CRUD operations.\newline
    5. Handle errors gracefully and manage high traffic scenarios."
\end{itemize}

\textbf{Analysis}

\begin{itemize}
    \item \textbf{Output 1}:
    \begin{itemize}
        \item \textbf{Strengths}: Provides a concise introduction with a clear step-by-step guide on starting a REST API project with Flask.
        \item \textbf{Weaknesses}: While it gives an initial direction, it lacks detailed implementation of each step, making it insufficient for beginners seeking a more complete understanding.
    \end{itemize}

    \item \textbf{Output 2}:
    \begin{itemize}
        \item \textbf{Strengths}: Offers a more detailed explanation, including specific code and a comprehensive example of a REST API structure.
        \item \textbf{Weaknesses}: Attempts to explain too many different concepts (e.g., authentication, session handling) without focusing on the core task of building a simple REST API, which could overwhelm beginners.
    \end{itemize}

    \item \textbf{Output 3}:
    \begin{itemize}
        \item \textbf{Strengths}: Breaks down the process systematically, covering environment setup and API structure with more technical depth.
        \item \textbf{Weaknesses}: Includes extraneous information such as Twilio integration and Elasticsearch, which may distract from the central task of building a REST API.
    \end{itemize}

    \item \textbf{Output 4}:
    \begin{itemize}
        \item \textbf{Strengths}: Provides an overview of the necessary steps for implementing a REST API with Flask and highlights key concepts such as CRUD operations and error handling.
        \item \textbf{Weaknesses}: Lacks specific code examples, making it harder for readers to translate the steps into an actual implementation.
    \end{itemize}
\end{itemize}

\textbf{Overall Analysis}

\begin{itemize}
    \item \textbf{Iteration 1} provides a basic step-by-step guide but lacks depth in key areas such as CRUD implementation and error handling.
    \item \textbf{Iteration 2} attempts to give a more complete explanation but introduces too many advanced concepts, which could confuse beginners.
    \item \textbf{Iteration 3} offers a detailed setup process but includes unnecessary information that distracts from the primary goal of building a REST API.
    \item \textbf{Iteration 4} gives a good overview but is too high-level and lacks concrete examples to help readers understand how to apply the steps.
\end{itemize}

\textbf{Conclusion}: \textbf{Iteration 2} offers the most comprehensive explanation but would benefit from focusing more on the key aspects of building a REST API, while \textbf{Iteration 1} is best suited for those seeking a simpler introduction.
\subsection{Case Study: Adding an Item to a Web Page's HTML List using JavaScript}

\textbf{Question}: 
\textit{"Write a JavaScript code snippet to add an item to a web page's HTML list."}

\textbf{Outputs}

\begin{itemize}
    \item \textbf{Output 1}: 
    "Sure, here is a simple JavaScript code snippet that will add an item with the text 'New Item' to your web page:\newline
    \texttt{// Select the first unordered list on the page (you can use this ID or class)\newline
    const ul = document.querySelector('ul');\newline
    if (ul) \{ \newline
    \ \ \ const li = document.createElement('li');\newline
    \ \ \ li.textContent = 'New Item';\newline
    \ \ \ ul.appendChild(li); \}}"

    \item \textbf{Output 2}: 
    "To add a new item to an HTML list using JavaScript, follow these steps:\newline
    1. Get the reference of the parent element where you want to insert the new list item. For example: \newline
    \texttt{const myList = document.getElementById('myList');}\newline
    2. Create a new list item node as a child of the selected parent element: \newline
    \texttt{const liNode = document.createElement('li');\newline
    liNode.innerHTML = 'New List Item';}\newline
    3. Append the newly created list item node to the specified parent container:\newline
    \texttt{myList.appendChild(liNode);}"

    \item \textbf{Output 3}: 
    "Here is a basic implementation of the functionality you described in JavaScript:\newline
    \texttt{// HTML structure with form and list\newline
    <form id='createForm'>\newline
    \ \ \ <label for='newItemText'>Enter new item text:</label><br>\newline
    \ \ \ <input type='text' id='newItemText'><br>\newline
    \ \ \ <button onclick='addToList()'>Add to list</button>\newline
    </form>\newline
    <ul id='existingListItems'></ul>\newline
    // JavaScript to handle the button click\newline
    function addToList() \{ \newline
    \ \ \ var inputText = document.getElementById('newItemText').value;\newline
    \ \ \ var li = document.createElement('li');\newline
    \ \ \ li.textContent = inputText;\newline
    \ \ \ document.getElementById('existingListItems').appendChild(li); \}}"

    \item \textbf{Output 4}: 
    "Here's an example of how you might implement this functionality using JavaScript:\newline
    \texttt{// HTML structure for input form and list\newline
    <div id='createForm'>\newline
    \ \ \ <label for='newItemText'>Enter new item text:</label><br>\newline
    \ \ \ <input type='text' id='newItemText'><br>\newline
    \ \ \ <button onclick='addToList()'>Add to list</button>\newline
    </div>\newline
    <ul id='existingListItems'></ul>\newline
    // JavaScript to handle the add button click\newline
    function addToList() \{ \newline
    \ \ \ var inputText = document.getElementById('newItemText').value;\newline
    \ \ \ var li = document.createElement('li');\newline
    \ \ \ li.textContent = inputText;\newline
    \ \ \ document.getElementById('existingListItems').appendChild(li); \}}"
\end{itemize}

\textbf{Analysis}

\begin{itemize}

    \item \textbf{Output 1}:
    \begin{itemize}
        \item \textbf{Strengths}: Provides a simple, concise code example for adding an item to an unordered list. The logic is easy to follow and useful for those who need a quick solution.
        \item \textbf{Weaknesses}: The example assumes the presence of an unordered list without explaining how the list is structured or created. It’s the simplest solution but lacks the depth and completeness of the other outputs.
    \end{itemize}

    \item \textbf{Output 2}:
    \begin{itemize}
        \item \textbf{Strengths}: Breaks the process into clear steps with explanations. The structured approach helps users follow along easily.
        \item \textbf{Weaknesses}: The code becomes slightly verbose and introduces unnecessary steps for what could be a simpler task. It also lacks the interactive element provided by Output 3 and Output 4.
    \end{itemize}

    \item \textbf{Output 3}:
    \begin{itemize}
        \item \textbf{Strengths}: Provides a complete implementation with both HTML and JavaScript, showcasing how to dynamically add items to an HTML list using a form. It demonstrates both code clarity and a practical use case, including event handling.
        \item \textbf{Weaknesses}: While comprehensive, the use of forms and event handling could add a small amount of complexity for users unfamiliar with those concepts.
    \end{itemize}

    \item \textbf{Output 4}:
    \begin{itemize}
        \item \textbf{Strengths}: Similar to Output 3, it provides a full implementation with HTML and JavaScript, with clean event handling and form interaction.
        \item \textbf{Weaknesses}: It includes form handling in a slightly clearer structure but doesn’t provide much additional value compared to Output 3. The form setup is slightly redundant for a simple task like adding a list item.
    \end{itemize}
\end{itemize}

\textbf{Overall Analysis}

\begin{itemize}
    \item \textbf{Iteration 1} is the most concise but lacks the depth and flexibility needed for a full implementation, making it the simplest but least powerful solution.
    \item \textbf{Iteration 2} is well-structured but lacks the interactive element of handling input dynamically through a form, making it less functional than Iteration 3 and 4.
    \item \textbf{Iteration 3} provides the most complete and functional solution, offering both HTML and JavaScript for a practical, interactive example.
    \item \textbf{Iteration 4} is also strong but is somewhat redundant compared to Iteration 3, offering slightly less detail in terms of implementation.
  \end{itemize}

\textbf{Conclusion}: \textbf{Iteration 3} offers the best balance between completeness and practical functionality. It provides a clear and interactive solution, making it the most useful for implementing the task of adding an item to an HTML list using JavaScript. 
\subsubsection{Case Study: Calculating the Average of a List in Python}

\textbf{Question}: 
\textit{"Write a Python function that takes a list of numbers and returns the average."}

\textbf{Outputs}

\begin{itemize}
    \item \textbf{Output 1}: 
    "Sure! Here is a simple Python function that does this:\newline
    \texttt{def calculate\_average(numbers):\newline
    \ \ \ total = sum(numbers)\newline
    \ \ \ return total / len(numbers)}"

    \item \textbf{Output 2}: 
    "Sure, here is a Python function called `calculate\_average` that takes a list of numbers as input and returns the average value. It also handles edge cases like an empty list:\newline
    \texttt{def calculate\_average(numbers):\newline
    \ \ \ return sum(numbers) / len(numbers) if len(numbers) > 0 else None}"

    \item \textbf{Output 3}: 
    "Here is a Python program that implements the `calculate\_average` function with additional checks and clarity:\newline
    \texttt{def calculate\_average(numbers):\newline
    \ \ \ if len(numbers) == 0:\newline
    \ \ \ \ \ \ return None\newline
    \ \ \ total = sum(numbers)\newline
    \ \ \ return total / len(numbers)}\newline
    This function includes a check for empty lists and provides a clean and understandable implementation with a return of `None` for edge cases."

    \item \textbf{Output 4}: 
    "Here is a modified version of the `calculate\_average` function that adds extra error handling and raises exceptions for certain cases:\newline
    \texttt{def calculate\_average(numbers):\newline
    \ \ \ if len(numbers) == 0:\newline
    \ \ \ \ \ \ raise ValueError('Input cannot be empty')\newline
    \ \ \ elif len(numbers) == 1:\newline
    \ \ \ \ \ \ return float('nan')\newline
    \ \ \ total = sum(numbers)\newline
    \ \ \ return total / len(numbers)}\newline
    This function handles both empty lists and single-element lists more strictly by raising exceptions and returning a NaN value where applicable."
\end{itemize}

\textbf{Analysis}

\begin{itemize}

    \item \textbf{Output 1}:
    \begin{itemize}
        \item \textbf{Strengths}: Provides the simplest and most direct solution for calculating the average of a list. It is easy to understand and quick to implement for basic use cases.
        \item \textbf{Weaknesses}: Does not handle any edge cases, such as empty lists, which may result in errors if used in real-world scenarios. It’s a good starting point but lacks robustness.
    \end{itemize}
    \item \textbf{Output 2}:
    \begin{itemize}
        \item \textbf{Strengths}: Provides a simple and practical solution with a basic error check for empty lists. The function is easy to understand and can handle the common case of an empty list by returning `None`.
        \item \textbf{Weaknesses}: The function only checks for empty lists but does not handle other potential issues such as single-element lists or non-numeric input. It is simpler than necessary for users looking for a more robust solution.
    \end{itemize}
\item \textbf{Output 3}:
\begin{itemize}
    \item \textbf{Strengths}: Provides a well-balanced solution with clear code and reasonable error handling. It accounts for edge cases such as empty lists and has a clean and readable structure. This output presents a practical and robust solution for calculating averages.
    \item \textbf{Weaknesses}: The implementation is straightforward, but it does not handle more complex exceptions such as non-numeric input, which could be useful for certain applications.
\end{itemize}

\item \textbf{Output 4}:
\begin{itemize}
    \item \textbf{Strengths}: Adds more advanced error handling by raising exceptions for empty lists and returning NaN for single-element lists. This output is ideal for users who want more strict error handling in specific edge cases.
    \item \textbf{Weaknesses}: While the function handles more complex scenarios, the added complexity may not be necessary for most average calculation tasks, making the function slightly over-engineered for basic purposes.
\end{itemize}

\end{itemize}
\textbf{Overall Analysis}

\begin{itemize}
   \item \textbf{Iteration 1} offers the simplest approach, but it doesn’t handle edge cases. It’s a good introductory solution but lacks robustness for more complex situations.
   \item \textbf{Iteration 2} is a simpler solution that handles empty lists but lacks more advanced error checking. It is useful for straightforward applications where minimal error handling is required.
   \item \textbf{Iteration 3} provides the most balanced solution, combining clarity with practical error handling. It is the best option for a well-rounded, everyday use case.
    \item \textbf{Iteration 4} introduces more strict error handling but adds complexity that may not be necessary for basic tasks. It’s useful for those who want more control over edge cases.
    
\end{itemize}

\textbf{Conclusion}: \textbf{Iteration 3} strikes the best balance between simplicity and practical error handling. It provides clear code with a clean solution for handling basic edge cases. \textbf{Iteration 4} is stronger in error handling but may be unnecessarily complex for most use cases. \textbf{Iteration 2} is simple and effective but lacks robustness, and \textbf{Iteration 1} is the most basic solution for introductory use.

\end{document}